\documentclass[%
 amsmath,amssymb,
superscriptaddress,
notitlepage
]{revtex4-1}

\usepackage[utf8]{inputenc} 
\usepackage[T1]{fontenc}    
\usepackage{hyperref}       
\usepackage{url}            
\usepackage{booktabs}       
\usepackage{amsfonts}       
\usepackage{nicefrac}       
\usepackage{microtype}      

\usepackage{physics}

\usepackage{graphicx}

\usepackage{mathtools}

\usepackage{comment}

\usepackage{enumitem}

\usepackage{amssymb}

\usepackage{algorithm}
\usepackage{algorithmicx}
\PassOptionsToPackage{noend}{algpseudocode}
\usepackage{algpseudocode}

\makeatletter
\let\OldStatex\Statex
\renewcommand{\Statex}[1][3]{%
  \setlength\@tempdima{\algorithmicindent}%
  \OldStatex\hskip\dimexpr#1\@tempdima\relax}
\makeatother

\algnewcommand{\LineComment}[1]{\State \(\triangleright\) #1}

\newcommand{\ipt}{\text{IP}_3}
\newcommand{\iptr}{\text{IP}_3\text{R}}
\newcommand{\nbf}{\boldsymbol{n}}

\newcommand{\pt}{\tilde{p}}
\newcommand{\thetabf}{\boldsymbol{\theta}}
\newcommand{\ubf}{\boldsymbol{u}}
\newcommand{\bbf}{\boldsymbol{b}}
\newcommand{\mubf}{\boldsymbol{\mu}}

\newcommand{\Fbf}{\boldsymbol{F}}
\newcommand{\fbf}{\boldsymbol{f}}

\newcommand{\cacytM}{\text{Ca}_\text{Cyt}}

\newcommand{\caerM}{\text{Ca}_\text{ER}}
\newcommand{\microMolar}{$\mu\text{M}$}
\newcommand{\microMolarSecInv}{(\microMolar $\times$ s)$^{-1}$}
\newcommand{\phibf}{\boldsymbol{\phi}}

\newcommand{\catp}{\text{Ca}^{2+}}
\newcommand{\catpCyt}{\text{Ca}_\text{Cyt}^{2+}}
\newcommand{\catpER}{\text{Ca}_\text{ER}^{2+}}
\newcommand{\concCyt}[1]{[#1]_\text{Cyt}}

\newcommand{\abf}{\boldsymbol{a}}

\newcommand{\mbf}{\boldsymbol{m}}
\newcommand{\vbf}{\boldsymbol{v}}

\newcommand{\zbf}{\boldsymbol{z}}

\newcommand{\figMain}[1]{Figure~#1 of the main text}
\newcommand{\figOne}{\figMain{1}}
\newcommand{\figTwo}{\figMain{2}}
\newcommand{\figThree}{\figMain{3}}
\newcommand{\figFour}{\figMain{4}}

\newcommand{\eqMain}[1]{Equation~(#1) of the main text}
\newcommand{\eqGaussClosure}{\eqMain{3}}
\newcommand{\eqPCA}{\eqMain{7}}
\newcommand{\eqML}{\eqMain{9}}
\newcommand{\eqTransHat}{\eqMain{10}}
\newcommand{\eqFourier}{\eqMain{13}}

\begin{document}

\title{
Physics-based machine learning for modeling stochastic IP3-dependent calcium dynamics
}

\author{Oliver K. Ernst}
\affiliation{
Department of Physics, University of California at San Diego, La Jolla, California
}
\affiliation{
Salk Institute for Biological Studies, La Jolla, California
}

\author{Tom Bartol}
\affiliation{
Salk Institute for Biological Studies, La Jolla, California
}

\author{Terrence Sejnowski}
\affiliation{
Salk Institute for Biological Studies, La Jolla, California
}
\affiliation{
Division of Biological Sciences, University of California at San Diego, La Jolla, California
}

\author{Eric Mjolsness}
\affiliation{
Departments of Computer Science and Mathematics, and Institute for Genomics and Bioinformatics, University of California at Irvine, Irvine, California
}

\date{\today}

\begin{abstract}
We present a machine learning method for model reduction which incorporates domain-specific physics through candidate functions. 
Our method estimates an effective probability distribution and differential equation model from stochastic simulations of a reaction network.
The close connection between reduced and fine scale descriptions allows approximations derived from the master equation to be introduced into the learning problem.
This representation is shown to improve generalization and allows a large reduction in network size for a classic model of inositol trisphosphate (IP3) dependent calcium oscillations in non-excitable cells.
\end{abstract}

\maketitle


\section{Introduction}


Modeling physical systems with machine learning is a growing research topic. Machine learning offers inference methods that can be computationally more efficient than first principles approaches, and that can generalize well from high dimensional datasets. Their successes in science span protein structure prediction~\cite{senior_2020} to solutions to the quantum many-body problem~\cite{carleo_2017}.

A key challenge is how to incorporate prior knowledge into the learning problem~\cite{osti_2019,mjolsness_2001,bezenac_2019}. 
This includes physical laws, symmetries and conservation laws.
For example, kernel methods have been improved by encoding symmetries~\cite{decoste_2002}, and convolutional neural networks (CNNs) have benefited from pose estimation~\cite{branson_2014}.
However, it remains difficult to introduce domain knowledge such as physical laws into machine learning.
Often, methods are used in a domain agnostic way~\cite{rupp_2012,racah_2017,giordani_2020,iten_2020}, in that physical processes are not introduced explicitly, but rather only implicitly present in the training data. For some applications this is an advantage~\cite{leemann_2019,magesan_2015}, but for scientific modeling it has at least three deficits. First, models can be challenging to train, having to internally rediscover already known function forms from large amounts of training data. Second, models can be difficult to interpret, requiring a large number of parameters to explain behavior that from first principles may be low dimensional. Third, the trained models may generalize poorly compared to approaches incorporating physical principles.

This paper introduces a method for modeling stochastic reaction networks that incorporates knowledge from the chemical master equation (CME) into the inference problem. 
This is made possible by representing the right hand side of a differential equation by a neural network~\cite{johnson_2015,ernst_2018,ernst_2019,raissi_2018,thiem_2020,long_2018}.
By using analytically derived approximations as inputs, the network is shown to improve generalization for a classic model of $\ipt$ dependent calcium oscillations~\cite{de_young_1992}. 
Additionally, reaction network conservation laws are incorporated into the framework.
From a subset of stochastic simulations, the trained model completes the full range of oscillations, and outperforms an equivalent domain-agnostic model.
The proposed approach is one avenue to improve machine learning for scientific modelling with domain-specific knowledge.


\subsection{Chemical kinetics at the fine scale}


\begin{figure*}[!ht]
	\centering
	\includegraphics[width=1.0\textwidth]{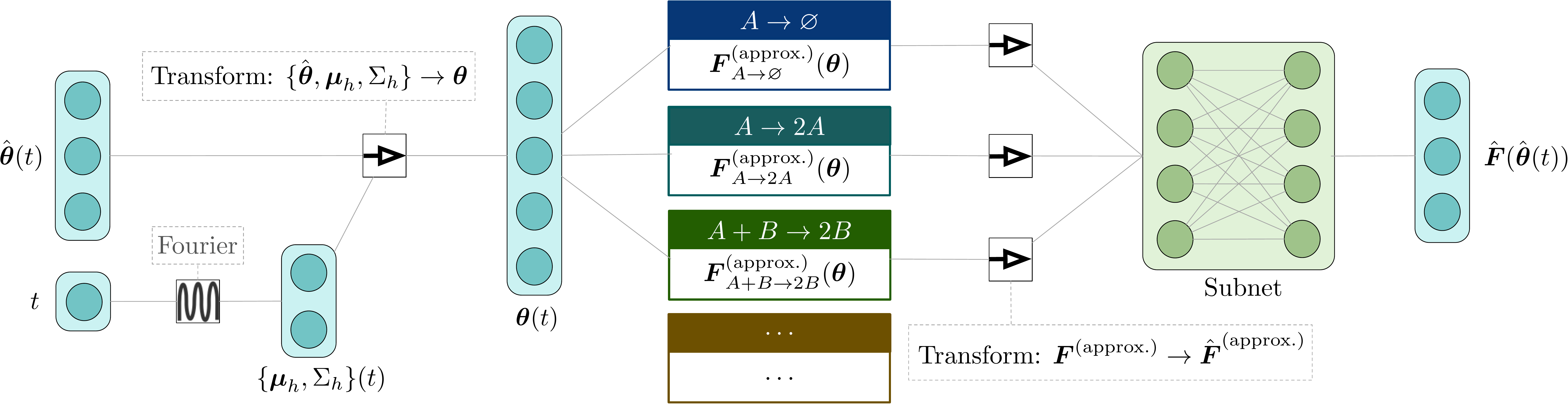}
	\caption{Architecture for the differential equation model~(\ref{eq:diffEqHat}), where the right hand side is parameterized by a neural network. Inputs and outputs of the subnet are also standardized.}
	\label{fig:1}
\end{figure*}

Consider a system described by the number of particles $\nbf = \{ n_A, n_B, \dots \}$ of species $\{ A, B, \dots \}$. The time evolution of the probability distribution over states $p(\nbf,t)$ is described by the CME:
\begin{equation}
\frac{d p(\nbf,t)}{dt} = \sum_{r=1}^R \sum_{\nbf^\prime} \left [ T_r(\nbf | \nbf^\prime) p(\nbf^\prime,t) - T_r(\nbf^\prime | \nbf) p(\nbf,t) \right ] ,
\label{eq:cme}
\end{equation}
where $T_r(\nbf | \nbf^\prime)$ is the propensity for the transition $\nbf^\prime$ to $\nbf$ under a reaction indexed by $r$. 

Only the simplest reaction networks are solvable exactly or perturbatively in the Doi-Peliti operator formalism~\cite{mattis_1998}. Further, the differential equations for moments derived from~(\ref{eq:cme}) generally do not close - equations for lower order moments depend on higher orders. For example, for $A + B \rightarrow 2B$:
\begin{equation}
\begin{split}
\frac{d \langle n_A \rangle}{dt} &\propto - \langle n_A n_B \rangle , \\
\frac{d \langle n_A n_B \rangle}{dt} &\propto - \langle n_A n_B^2 \rangle + \langle n_A^2 n_B \rangle - \langle n_A n_B \rangle .
\end{split}
\label{eq:momsEqsExample}
\end{equation}
This infinite hierarchy requires a moment closure approximation, such as the Gaussian closure approximation:
\begin{equation}
p(\nbf,t) \sim {\cal N}(\nbf | \mubf_p(t), \Sigma_p(t)) ,
\label{eq:gaussClosure}
\end{equation}
where $\mubf_p,\Sigma_p$ are the mean and covariance under $p$ at an instant in time. In practice, it is challenging to choose the optimal closure approximation, since it is not clear which higher order moments will become relevant over long times.

Alternatively, the Gillespie algorithm~\cite{gillespie_1977} can be used to simulate stochastic trajectories of reaction networks. This is popular in biology~\cite{mcell_2,bartol_2015}, at the cost of computation time for collecting sufficient statistics.
Motivated by data-driven methods, we next propose a framework to learn closure approximations from stochastic simulations.


\section{Physics-based machine learning}



\subsection{Reduced model}


We seek a reduced model that can be trained on stochastic simulations, but also incorporates physical knowledge to improve generalization. This connection can be made by a dynamic Boltzmann distribution (DBD)~\cite{ernst_2018,ernst_2019,ernst_2019_arxiv}, consisting of an effective probability distribution with time-dependent interactions $\thetabf(t)$ in the energy function:
\begin{equation}
\pt(\nbf; \thetabf(t)) = \frac{1}{Z(t)} \exp[ - E(\nbf; \thetabf(t)) ]
,
\label{eq:dbd}
\end{equation}
and a differential equation system for the parameters:
\begin{equation}
\frac{d \thetabf(t)}{dt} = \Fbf(\thetabf(t); \ubf)
,
\label{eq:diffEq}
\end{equation}
for some functions $\Fbf$ with parameters $\ubf$, with a given initial condition $\thetabf(t=0) = \thetabf_0$. 
The Boltzmann distribution \textit{ansatz} is motivated by the connection to graphical models~\cite{johnson_2015}.
In this work, the reduced model~(\ref{eq:dbd}) considered is that of probabilistic principal component analysis (PCA), a popular choice for dimensionality reduction~\cite{bishop_2006}. The parameters in the energy function are:
\begin{equation}
\thetabf(t) = \{ \bbf, W, \sigma^2, \mubf_h, \Sigma_h \} (t) ,
\end{equation}
and the distribution is Gaussian:
\begin{equation}
\begin{split}
\pt(\nbf; \thetabf(t)) &= {\cal N}(\nbf | \mubf(t), C(t)) , \\
\mubf(t) &= \begin{pmatrix}
\bbf + W \mubf_h \\
\mubf_h
\end{pmatrix}(t) , \\
C(t) &= \begin{pmatrix}
W W^\intercal + \sigma^2 I & W \Sigma_h \\
\Sigma_h W^\intercal & \Sigma_h
\end{pmatrix}(t) .
\end{split}
\label{eq:pca}
\end{equation}
Splitting the species into visible $\nbf_v$ of size $N_v$ and hidden $\nbf_h$ of size $N_h$ gives the more familiar form:
\begin{equation}
\begin{split}
\pt(\nbf_h ; \thetabf(t)) &= {\cal N} ( \nbf_h | \mubf_h, \Sigma_h ) , \\
\pt(\nbf_v | \nbf_h ; \thetabf(t)) &= {\cal N} (\nbf_v | \bbf + W (\mubf_h + \nbf_h), \sigma^2 I ) .
\end{split}
\end{equation}


\subsection{Maximum likelihood at an instant in time}


At an instant in time, $\mubf_h$ and $\Sigma_h$ are arbitrary; across time, the differential equation~(\ref{eq:diffEq}) depends on these variables. For $\mubf_h = \boldsymbol{0}$ and $\Sigma_h = I$, the maximum likelihood (ML) solution is:
\begin{equation}
\begin{split}
\hat{W}_\text{ML}(t) &= U_q(t) (L_q(t) - \sigma_\text{ML}^2(t) I)^{1/2} R , \\
\sigma_\text{ML}^2(t) &= \frac{1}{N_v - q} \sum_{i=q+1}^{N_v} \lambda_i(t) , \\
\hat{\bbf}_\text{ML}(t) &= \frac{1}{M} \sum_{i=1}^M X_i(t) ,
\end{split}
\label{eq:ml}
\end{equation}
where $M$ is the number of samples, $X(t)$ is the data matrix of size $M \times N_v$, and $U_q(t)$ and $L_q(t)$ are the normalized eigenvectors and eigenvalues of the data covariance matrix for the $1 \leq q \leq N_v$ largest eigenvalues. $R$ is a rotation matrix that can be taken as $R=I$. The transformation to arbitrary $\mubf_h, \Sigma_h$ is:
\begin{equation}
\begin{split}
\bbf_\text{ML}(t) &= \hat{\bbf}_\text{ML}(t) - \hat{W}_\text{ML}(t) \Sigma_h^{-1/2}(t) \mubf_h(t) , \\
W_\text{ML}(t) &= \hat{W}_\text{ML}(t) \Sigma_h^{-1/2}(t) .
\end{split}
\label{eq:transHat}
\end{equation}
with matrix square root as $(A^{1/2})^\intercal A^{1/2} = A$. For convenience, let $\hat{\thetabf}(t) = \{ \hat{\bbf}, \hat{W}, \sigma^2 \} (t)$ denote the \textit{standard parameters}.


\subsection{Linking snapshots in time}


Given a set of training data, the ML parameters $\thetabf_\text{ML}(t)$ can be obtained at each timepoint. 
To link snapshots in time, the form of the differential equations~(\ref{eq:diffEq}) must be chosen.
The known CME physics is used to guide this choice by 
deriving an approximation $\Fbf^\text{(approx.)}$ to the true time evolution $\Fbf$ as follows.

At any point in time, the distribution defined by $\thetabf(t)$ has observables $\phibf(t) = \{ \mubf, C \}(t)$. For a single reaction like $A + B \rightarrow 2 B$, these evolve as $d \phibf_{A+ B \rightarrow 2B} / dt$ according to a hierarchy of moments like~(\ref{eq:momsEqsExample}), derived from the CME. Under the Gaussian closure approximation~(\ref{eq:gaussClosure}), the equations for the moments are closed $d \phibf_{A+ B \rightarrow 2B} / dt \sim d \phibf_{A+ B \rightarrow 2B}^\text{(closed)} / dt$. To convert back to the parameter frame, only some observables are tracked exactly. While arbitrary, the natural choice is $d \{ \mubf_v, C_{vh}, \text{Tr}(C_v), \mubf_h, \Sigma_h \} / dt$ which match the dimensions of $\thetabf$. The equations corresponding to this conversion $d \phibf_{A+ B \rightarrow 2B}^\text{(closed)} / dt \rightarrow d \thetabf_{A+B \rightarrow 2B}^\text{(closed)} / dt$ are obtained by differentiating~(\ref{eq:pca}). The result $\Fbf_{A+B \rightarrow 2 B}^\text{(approx.)} \equiv d \thetabf_{A+B \rightarrow 2B}^\text{(closed)} / dt$ is an approximation to the time evolution of $\thetabf(t)$ under this reaction~(Supplemental material).

By considering a variety of reaction processes in this manner, a set of candidates was generated and used to parameterize the differential equations~(\ref{eq:diffEq}).
It has been shown that the linearity of the CME in reactions extends to this form of the reduced model~\cite{ernst_2018}.
However, a linear model for~(\ref{eq:diffEq}) generalizes poorly when the data is not well-represented by a sparse set of available candidates~\cite{brunton_2016}.

Instead, let the right hand side of the differential equations~(\ref{eq:diffEq}) be given by a neural network with a special architecture shown in Figure~\ref{fig:1}. At each point in time with standard parameters $\hat{\thetabf}(t)$, the inputs are the different reaction approximations. The outputs are the derivatives:
\begin{equation}
\frac{d \hat{\thetabf}(t) }{ dt } = \hat{\Fbf}(\hat{\thetabf}(t) ; \ubf) ,
\label{eq:diffEqHat}
\end{equation}
where the parameters $\ubf$ are those of the neural network. The model is trained to optimize the $L_2$ loss:
\begin{equation}
S = \sum_{t=1}^T \left ( 
\frac{d \hat{\thetabf}_\text{ML} (t)}{dt}
- \hat{\Fbf}( \hat{\thetabf}_\text{ML}(t) ; \ubf ) 
\right )^2 .
\end{equation}
The training data is obtained from the ML parameters $\hat{\thetabf}_\text{ML}(t)$ for $t=1,\dots,T$ by first computing $d \hat{\thetabf}_\text{ML}(t) / dt$ using total variation regularization~\cite{xiao_2011} to differentiate the noisy signals.
After training, the integration of~(\ref{eq:diffEqHat}) is stable if the Jacobian of the candidates $\partial \hat{\Fbf}_{A+B \rightarrow 2B}^\text{(approx.)} / \partial \hat{\thetabf}$ is small. To reduce the Jacobian, the data matrices $X(t)$ are transformed using a standardizing transformation~(Supplemental material).

In principle, the standard parameters $\hat{\thetabf}$ where $\mubf_h=\boldsymbol{0}, \Sigma_h = I$ can be used to calculate the reaction approximations.
Instead, to improve generalization, the latent parameters $\mubf_h,\Sigma_h$ are learned as a Fourier series. For a fixed set of $L$ frequencies $\fbf$, let $\Sigma_h$ be diagonal and let:
\begin{equation}
\begin{split}
\mu_{h,i}(t) &= s ( \abf^{(\mu,i)}, \bbf^{(\mu,i)} ) , \\
\Sigma_{h,i,i}(t) &= 1 + \epsilon + s ( \abf^{(\Sigma,i,i)}, \bbf^{(\Sigma,i,i)} ) , \\
s(\abf,\bbf) &=
\frac {
\sum_{l=1}^L \left (
a_l \cos ( f_l t )
+ b_l \sin ( f_l t )
\right )
}{
\text{max} \left ( \sum_{l=1}^L (|a_l| + |b_l|), 1 \right )
+ \epsilon
} ,
\end{split}
\label{eq:fourier}
\end{equation}
where $\epsilon$ is small and coefficients $\abf,\bbf$ are learned. This lets $\mubf_h$ oscillate in $[\boldsymbol{-1},\boldsymbol{1}]$ and $\Sigma_h$ around the identity. Finally, since $\mubf_h,\Sigma_h$ are unknown from the data, the approximations $\Fbf_\text{reaction}^\text{(approx.)}$ are converted back to the standard space $\hat{\Fbf}_\text{reaction}^\text{(approx.)}$ using~(\ref{eq:transHat}).


\section{IP3 dependent calcium oscillations}


\begin{figure}[t]
	\centering
	\includegraphics[width=0.7\linewidth]{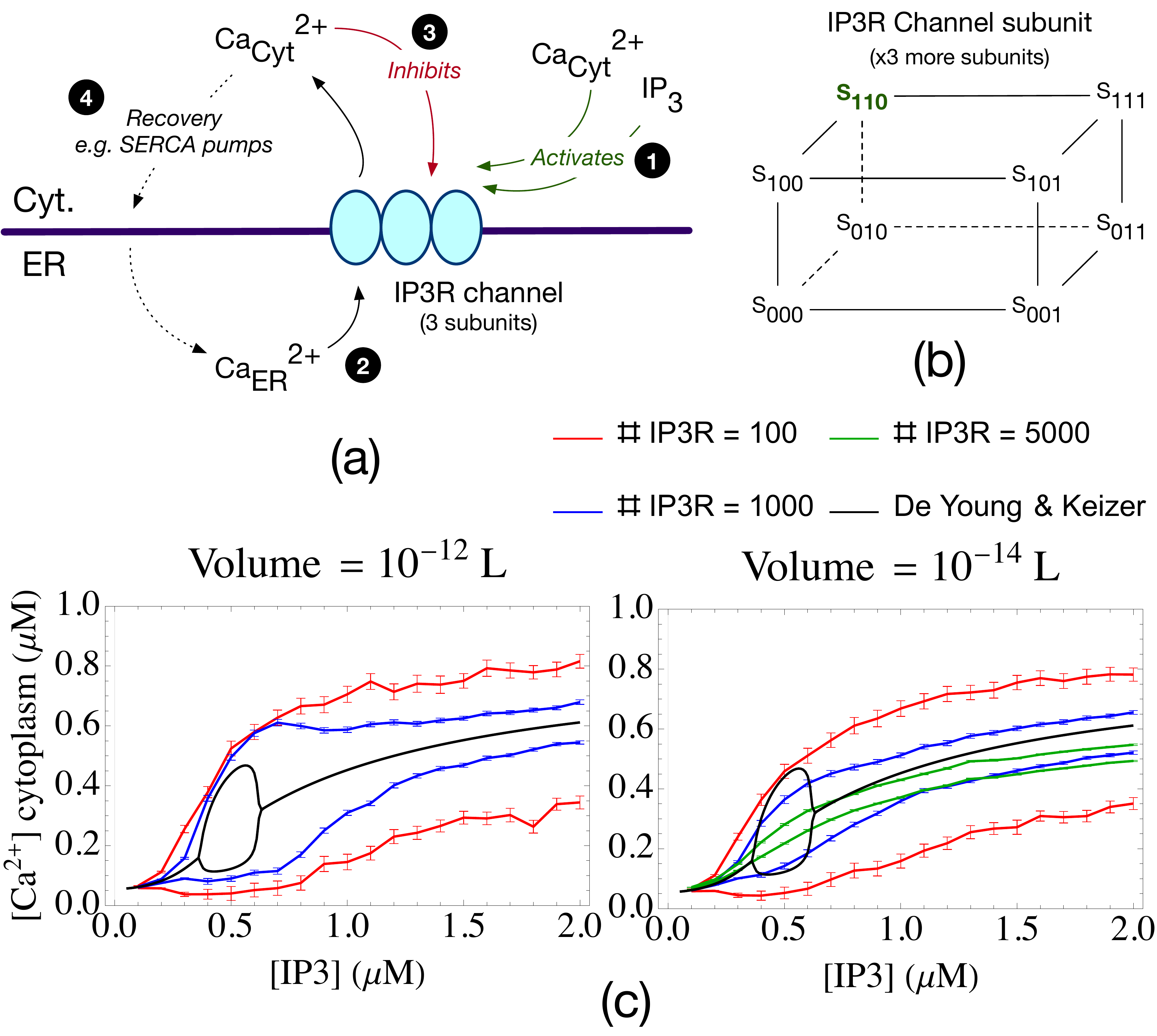}
	\caption{
	(a) Schematic of $\ipt$ dependent calcium oscillations. Clusters of $\iptr$s in the ER membrane are activated by cytosolic calcium and $\ipt$ (1), allowing calcium transport into the cytoplasm (2). Further binding inhibits the channel (3), and eventual recovery recycles the calcium store (4).
	(b) Channel states of one of three subunits~(Supplemental material).
	(c) Range of oscillations in the stochastic and deterministic~\cite{de_young_1992} models for different volumes and numbers of $\iptr$s. Error bars indicate $95\%$ confidence levels. 
	}
	\label{fig:2}
\end{figure}

The proposed physics-based ML method is demonstrated for calcium oscillations in non-excitable cells~\cite{voorsluijs_2019}.
These occur due to calcium influx into the cytoplasm from stores in the endoplasmic reticulum (ER) through $\ipt$ receptors ($\iptr$s) in the membrane. A classic model by~\citet{de_young_1992} uses ordinary differential equations and treats the channel at equilibrium, as shown in Figures~\ref{fig:2}. 

A key result is a bifurcation diagram for calcium oscillations, shown in Figure~\ref{fig:2}(c). A Hopf bifurcation occurs at $[\ipt] \sim 0.4 \mu$M beyond which oscillations arise. Beyond $[\ipt] \sim 0.6 \mu$M, a stable elevated level of calcium is observed.

Figure~\ref{fig:2}(c) compares the bifurcation diagram with the range of oscillations observed in a stochastic version of the~\citet{de_young_1992} model. The receptor channel states and transport through the channel are simulated using the Gillespie method, with identical parameters to those in~\cite{de_young_1992}.
For the stochastic model, two cytoplasm volumes are considered: $10^{-12}$~L and $10^{-14}$~L, and the number of $\iptr$ is varied.
The range of oscillations show the maximum/minimum over $40$~s of the mean calcium concentration plus/minus a standard deviation. Spontaneous calcium spikes continue to arise in the stochastic model even at high $\ipt$ concentrations.

\begin{figure}[!ht]
	\centering
	\includegraphics[width=0.7\linewidth]{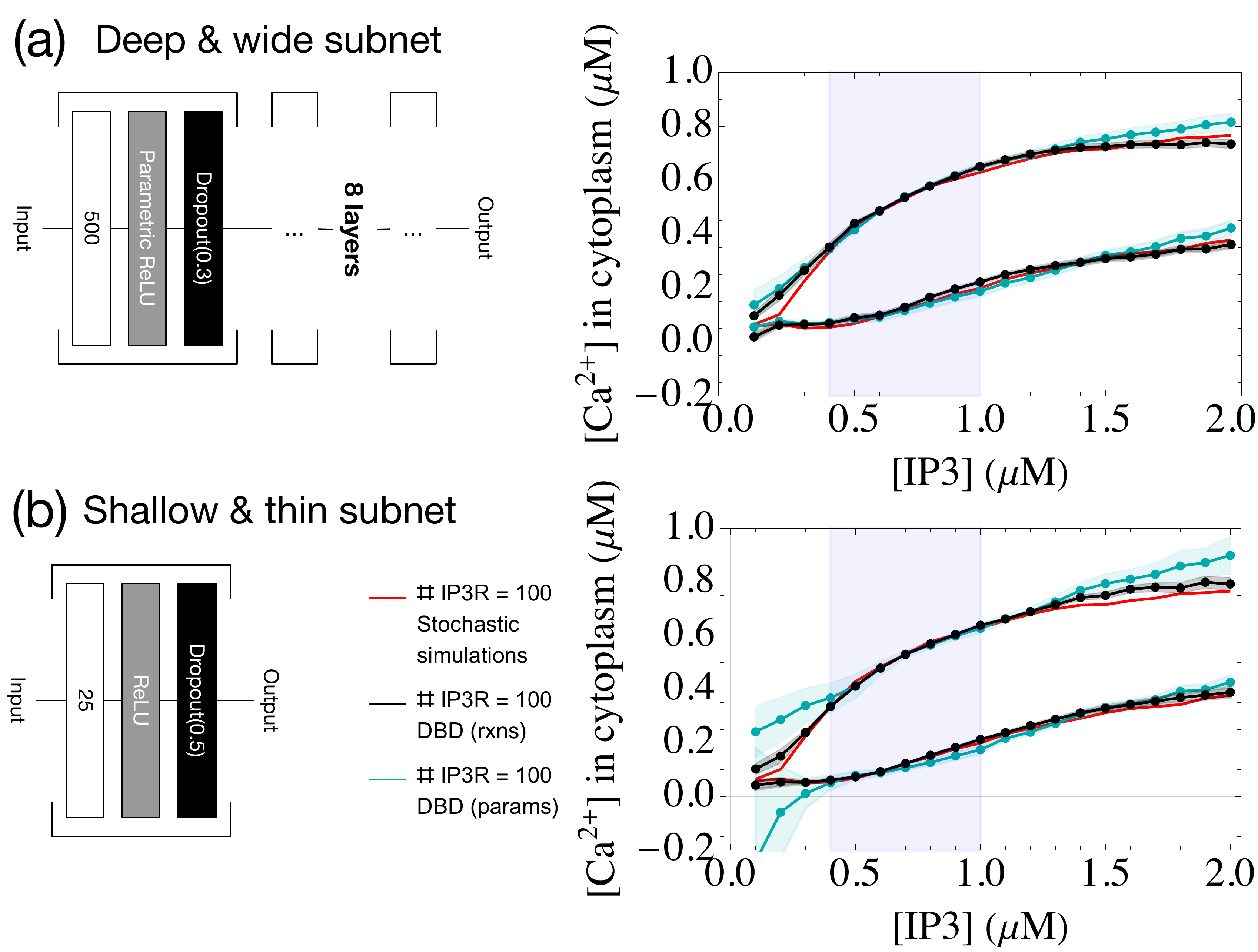}
	\caption{
	Incorporating reaction approximations improves generalization (black).
	Two subnets are compared: (a) a deep \& wide subnet, and (b) a shallow \& thin subnet. 
	The $\ipt$ concentrations used as training data are highlighted (blue). A comparison architecture using the same subnet but missing reaction approximations is also shown (cyan). Shading shows $95\%$ confidence intervals from $10$ optimization trials.
	}
	\label{fig:3}
\end{figure}


\subsection{Learning calcium oscillations}


The DBD architecture is applied to learn calcium oscillations over a subset of $\ipt$ concentrations~(all code is available online~\cite{code}).
Figure~\ref{fig:3} shows the range of oscillations learned for $V=10^{-14}$L and $100$~$\iptr$ receptors. The training data consists of simulations at $\ipt$ concentrations over $[0.4,1] \mu$M in intervals of $0.1 \mu$M. Two subnet models are explored: a deep \& wide subnet consisting of $8$ layers of width $500$ units, and a shallow \& thin subnet consisting of a single layer of $25$ units, both using ReLU activation functions and dropout. Three species are used in the effective probability distribution~(\ref{eq:dbd}): $\catp,\ipt$ and a latent species $X$. The reaction approximations used are those from enumerating the Lotka-Volterra system~(Supplemental material): $P \rightarrow 2P$, $H \rightarrow \varnothing$, and $H+P \rightarrow 2H$, allowing each combination of $\{ H,P \}$ from $\{ \catp,\ipt,X \}$.

To demonstrate how domain-specific knowledge improves generalization, a comparison \textit{parameter-only model} is shown, equivalent to Figure~\ref{fig:1} but missing the reaction approximations~(Supplemental material). Both models are trained using the Adam optimizer~\cite{kingma_2014} with batch size~$64$ and learning rate~$10^{-3}$. The deep subnet is trained for~$25$ rounds with weight clipping beyond a cutoff magnitude of~$5$; the shallow subnet for~$200$ rounds and weight cutoff~$1$.
Between the parameter-only and the \textit{reaction model}, the latter generalizes better to $\ipt$ concentrations not observed during training. Further, the reaction model outperforms the comparison at keeping concentrations non-negative over the domain explored, although this is not explicitly enforced.

\begin{figure}[!ht]
	\centering
	\includegraphics[width=0.7\linewidth]{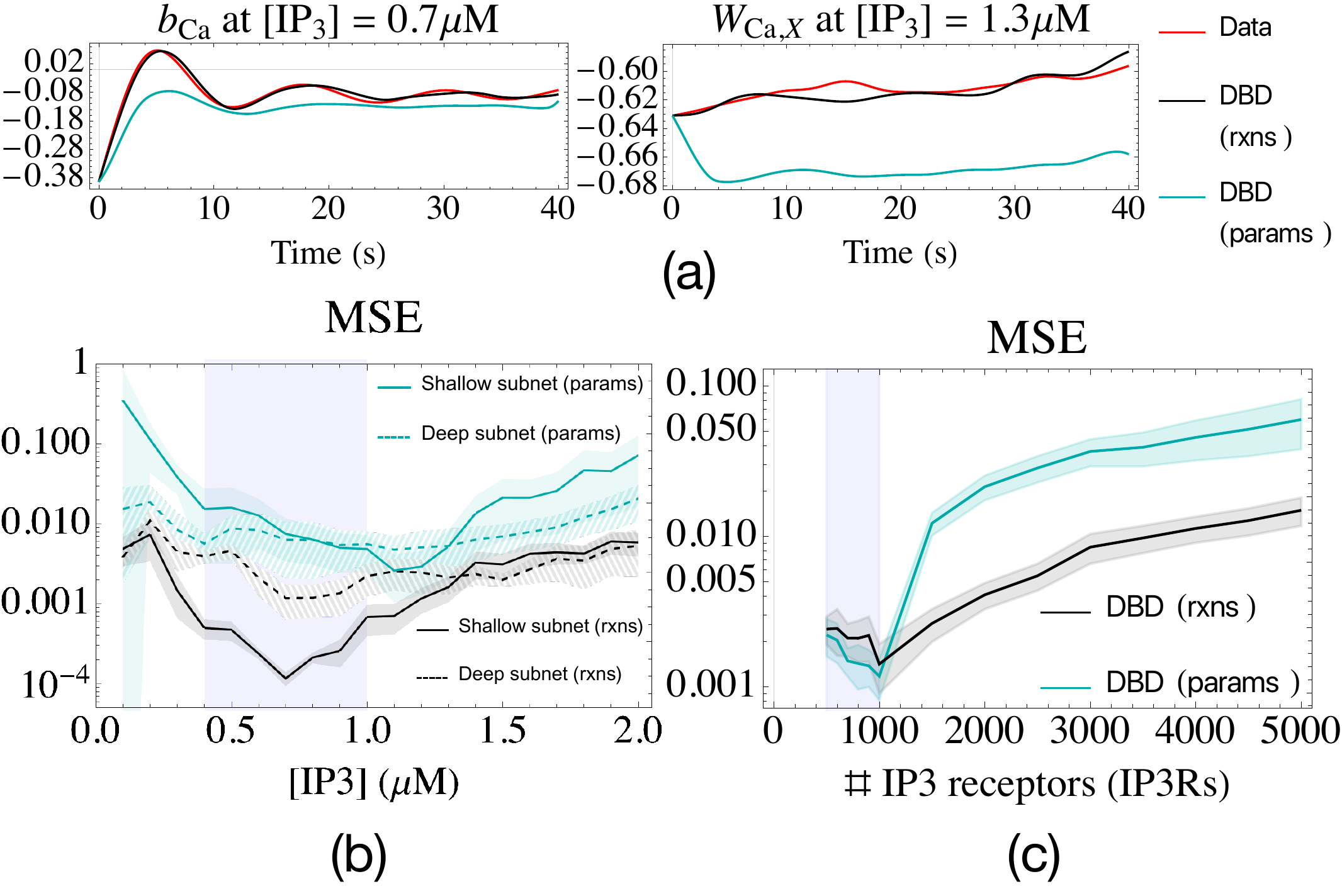}
	\caption{
	(a) Parameters from the shallow subnet model from slices of Figure~\ref{fig:3} at $[\ipt]=0.7 \mu$M and $[\ipt]=1.3 \mu$M. The reaction model learns a more detailed model.
	(b) MSE for the learned parameters $\hat{\thetabf}$ in the models of Figure~\ref{fig:3}, with training data in blue. Models with reaction approximations decrease MSE by up to an order of magnitude. Shading shows $95\%$ confidence intervals from $10$ optimization trials.
	(c) MSE of a second model generalizing in $\ipt$ receptor number, with $95\%$ confidence intervals from $40$ trials, and training data in blue.
	}
	\label{fig:4}
\end{figure}

The generalization of the parameter-only model is better for the deep subnet than for the shallow subnet, partly because multiple layers of dropout improve generalization. However, the reaction model generalizes well even for the very low parameter shallow subnet.
Figure~\ref{fig:4}(a) shows the integrated parameters at two slices of $\ipt$ using the Euler method. The reaction models learn the curves more exactly on both training and validation sets. This is quantified by a lower mean-squared error (MSE) shown in Figure~\ref{fig:4}(b).

\subsection{Encoding conservation}


A second axis to generalize in is the number of $\iptr$s. The PCA model is now formulated for four species $\{ \catp, \ipt, \iptr, X \}$ (the variance of $\iptr$ is set to a small constant $10^{-7}$ in the ML step). Since the receptor number is conserved in the simulations, the reaction approximations are extended with three of the form $A + \iptr \rightarrow \iptr$, where $A$ is one of $\{ \catp, \ipt, X \}$. This explicitly conserves $\iptr$ in the input approximations.

Figure~\ref{fig:4}(c) shows the MSE over parameters for this model, trained on simulations at $\iptr$s over~$[500,1000]$ in intervals of~$100$. The reaction model outperforms the parameter-only model over the validation set covering~$1000$ to~$5000$ receptors.
Training used the Adam optimizer for~$25$ rounds, with learning rate~$10^{-3}$ and batch size~$64$.
The subnet has~$5$ hidden layers of~$150$ units, ReLU activations, dropout rate~$0.1$, and weight cutoff~$0.5$.


\section{Discussion}


The power of DBDs is that knowledge of the domain can be explicitly built into the learning problem. This is possible due to the tight connection between reduced and fine scale models. Both are Markovian, depending only on the current point in parameter space. Moreover, because the reduced model is formulated by differential equations~(\ref{eq:diffEq}), reaction network physics could be built in through candidate functions derived from the master equation. Additionally, conservation laws for $\iptr$ were included in the network inputs. 
These connections to the underlying physics differentiate DBDs from how neural networks are commonly used for time series regression, and other methods including hidden Markov models (HMMs) and recurrent neural networks (RNNs). 
A further desired property is that the learned covariance matrix $C(t)$ is positive semidefinite at all times. This is the case for the PCA model~(\ref{eq:pca}) due to the transformation~(\ref{eq:transHat}), but not for a generic Gaussian distribution.
Additionally, the means $\mubf(t)$ should be non-negative to represent particle counts. This is not enforced explicitly, but is observed for the reaction models in Figure~\ref{fig:3}.

Other methods have been proposed that learn a neural network representing a differential equation directly from parameters without explicitly incorporating domain-specific physics~\cite{raissi_2018,thiem_2020,long_2018}. A related method that uses candidates is SINDy~\cite{brunton_2016}, but its differential equations are linear and struggle with model reduction, where candidates do not include the true dynamics. Further, its candidates are arbitrary polynomial forms, and not necessarily connected to underlying physics. For graphical models, graph-constrained correlation dynamics (GCCD)~\cite{johnson_2015} has used polynomial and exponential candidates non-linearly with neural networks. Parameterizations using basis functions from finite elements~\cite{ernst_2019,ernst_2019_arxiv} have also been used. In these cases, for graphical models other than PCA~(\ref{eq:pca}), the ML parameters can be estimated by the Boltzmann machine learning algorithm~\cite{ackley_1985} or by expectation maximization~\cite{bishop_2006}.

One avenue for improvement is to include approximations for small networks rather than just individual reactions. DBDs may also be extendable to delay differential equations to improve regression performance. Alternatively, this may be implemented using tailored input reaction motifs. Further closure approximations beyond the Gaussian closure can also be included as candidates. 

While the models considered have no spatial dependence, the approach is equally valid for spatial systems~\cite{ernst_2018,ernst_2019}. A spatial model of $\ipt$-dependent calcium oscillations may include plasma membrane pumps and feedback on $\ipt$ production. One application of DBDs is to synaptic neuroscience, where simulations of signaling pathways~\cite{bartol_2015} could be used to build models that are computationally efficient and generalize well to new stimulation patterns. Beyond reaction-diffusion systems, applications to other domains such as neural populations~\cite{ohira_1997} may be possible.


\begin{acknowledgements}
This work was supported by NIH R56-AG059602 (E.M., O.K.E., T.M.B., and T.J.S.), NIH P41-GM103712, NIH R01-MH115556 (O.K.E., T.M.B., and T.J.S.), Human Frontiers Science Program Grant No. HFSP-RGP0023/2018, the UC Irvine Donald Bren School of Information and Computer Sciences, NSF Grant No. PHY-1748958, NIH Grant No. R25GM067110, and the Gordon and Betty Moore Foundation Grant No. 2919.02 (E.M.).
\end{acknowledgements}


\bibliographystyle{plainnat_custom}
\bibliography{bibliography}


\appendix


\section{Code}


All code is available as part of this supplemental material as well as online~\cite{code}. This includes codes for the stochastic simulations, learning problems, and notebooks to reproduce figures. See the ``Readme" files included with the code for directions.


\section{IP3 dependent calcium oscillations}



\subsection{Stochastic models}
\label{app:stochSims}

\begin{figure}[htp]
	\centering
	\includegraphics[width=0.5\textwidth]{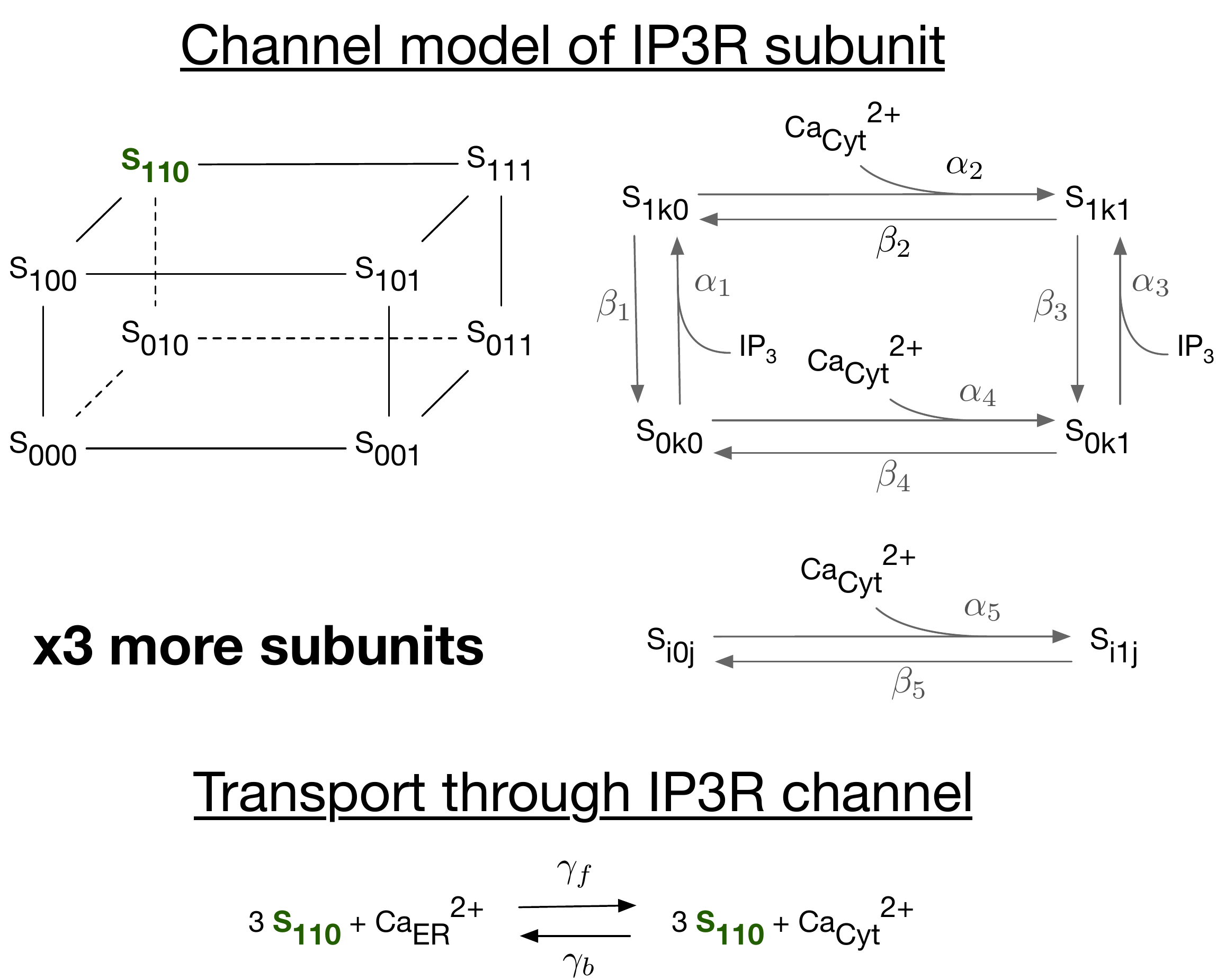}
	\caption{Channel models for the $\iptr$ and transport through the channel, which is open when one calcium ion and one $\ipt$ are bound. A further calcium binding inhibits the channel. Peak conductance is observed when three subunits are open; a fourth that is physically observed is not modelled. The open state is $S_{110}$, indicated in green.}
	\label{fig:channels}
\end{figure}

\figTwo shows a schematic of the model of $\ipt$ dependent calcium oscillations in non-excitable cells. Clusters of $\ipt$ receptors ($\iptr$s) in the membrane of the endoplasmic reticulum are activated by cytosolic calcium and $\ipt$, allowing transport through the channel into the cytoplasm. The channel model for a single $\ipt$ receptor subunit is shown in Figure~\ref{fig:channels}. While the receptor is known to be composed of four subunits, the peak conductance is observed when only three are open. Hence, the original model of~\citet{de_young_1992} considers only three subunits. The reactions in the receptor subunit are:
\begin{equation}
\begin{split}
S_{1k0} + \cacytM &\overset{\alpha_2}{\underset{\beta_2}{\rightleftharpoons}} S_{1k1} , \\
S_{0k0} + \cacytM &\overset{\alpha_4}{\underset{\beta_4}{\rightleftharpoons}} S_{0k1} , \\
S_{0k0} + \ipt &\overset{\alpha_1}{\underset{\beta_1}{\rightleftharpoons}} S_{1k0} , \\
S_{0k1} + \ipt &\overset{\alpha_3}{\underset{\beta_3}{\rightleftharpoons}} S_{1k1} , \\
S_{i0j} + \cacytM &\overset{\alpha_5}{\underset{\beta_5}{\rightleftharpoons}} S_{i1j} ,
\end{split}
\label{eq:stateRxns}
\end{equation}
where the open state is $S_{110}$. Table~\ref{tab:table1} gives the parameter values used for stochastic simulations, which are the same values used in the original~\citet{de_young_1992} model. The molecular-based reaction rates are obtained from the concentration-based rates as:
\begin{equation}
\begin{split}
\alpha_i &= \frac{a_i}{c_A \times V_\text{Cyt}} , \\
\beta_i &= b_i ,
\end{split}
\end{equation}
as derived in Section~\ref{sec:convert}, where $c_A$ is Avogadro's constant and $V_\text{Cyt}$ is the volume of the cytoplasm.

Transport through the channel is given by the reactions:
\begin{equation}
3 S_{110} + \caerM \overset{\gamma_f}{\underset{\gamma_b}{\rightleftharpoons}} 3 S_{110} + \cacytM ,
\end{equation}
as also derived in Section~\ref{sec:convert} from the differential equation model by~\citet{de_young_1992}.

The recovery of calcium from the cytoplasm is attributed to ATP-driven pumps such as SERCA pumps. In this model, since the density of these pumps is as high as $1000/\mu\text{m}^2$~\cite{bartol_2015} and the ER is a highly folded structure with a large surface area, this process is not modeled using stochastic particle-based methods, but rather by differential equations:
\begin{equation}
\begin{split}
\frac{d [\cacytM]}{dt} &= J_1 - J_2 , \\
J_1 &= c_1 v_2 ( c_1^{-1} [\caerM] - [\cacytM] ) , \\
J_2 &= \frac{v_3 [\cacytM]^2 }{ [\cacytM]^2 + k_3^2 } ,
\end{split}
\label{eq:currents}
\end{equation}
where $J_1$ is a leak current, and $J_2$ is the ATP-driven recovery of calcium back to the ER. See also Section~\ref{sec:convert}.

\begin{figure*}[htp]
	\centering
	\includegraphics[width=1.0\textwidth]{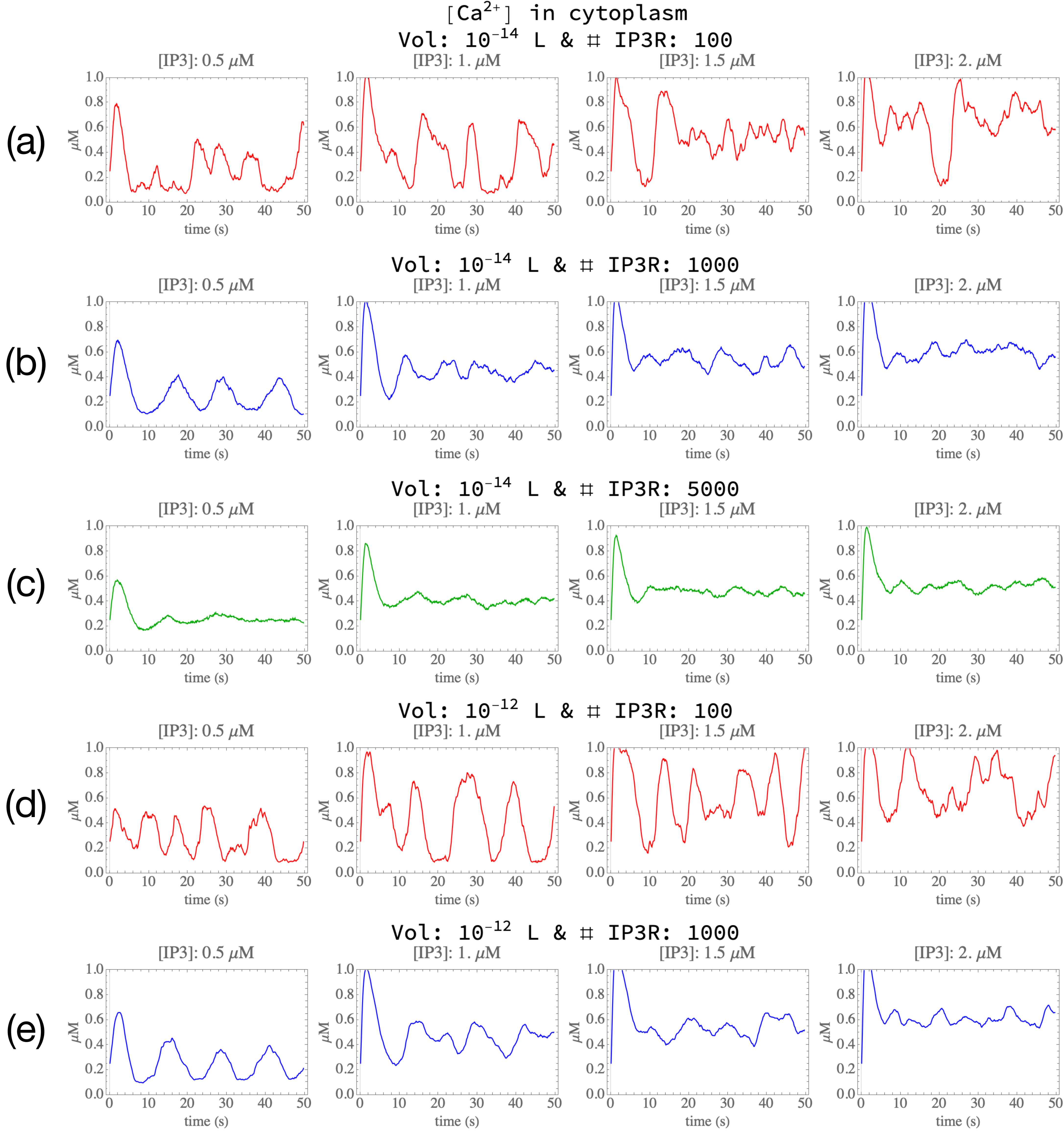}
	\caption{Stochastic simulations of cytosolic calcium oscillations for (a) $100$ $\iptr$ and (b) $1000$ $\iptr$ at various $\ipt$ concentrations and cytoplasm volume of $10^{-14}$L. Calcium spikes are observed at all concentrations, and are more pronounced at lower receptor number.}
	\label{fig:appSS}
\end{figure*}

Examples of the stochastic simulations are shown in Figure~\ref{fig:appSS}. The initial number of $\catp$ and $\ipt$ are sampled from the Gaussian distributions:
\begin{equation}
\begin{split}
[\cacytM]_0 &= {\cal N}( \mu_0([\cacytM]), \sigma_0^2([\cacytM]) ) , \\
[\ipt]_0 &= {\cal N}( \mu_0([\ipt]), \sigma_0^2([\ipt]) ) ,
\end{split}
\label{eq:initConcs}
\end{equation}
where the parameter values are given in Table~\ref{tab:table1}.

After sampling the initial counts, an initial simulation is used to initialize the states of the $\ipt$ receptors. Let the numbers of particles corresponding to the concentrations~(\ref{eq:initConcs}) be $n_{\cacytM,0}$ and $n_{\ipt,0}$, from which the number of calcium particles in the ER $n_{\caerM,0}$ can be calculated using $c_0,c_1$. The $\iptr$ are initialized to the specified and fixed number $n_{\iptr}$ of receptors, all in state $S_{000}$, and all other receptor states $S_{ijk}$ with population zero. The initial simulation is run with only the $\iptr$ state reaction system~(\ref{eq:stateRxns}), and where the number of $\cacytM, \ipt, \caerM$ is conserved, i.e. fixed to their initial values. The duration of the initial simulation is $10$~s. From this, the initial states of the receptors are taken for the main simulation as the average state values over the last $4$~s of the initial simulation.

The main simulations are run from $t=0$ to $t=T_\text{max}$, with the count of each species written out at intervals $\Delta t^\text{(write)}$. The currents from the differential equations~(\ref{eq:currents}) are updated at short time intervals $\Delta t^\text{(diff. eq.)}$, with all parameters as given in Table~\ref{tab:table1}.

\begin{table*}[htp]
\caption{\label{tab:table1}%
Parameter values used for stochastic simulations.
}
\begin{ruledtabular}
\begin{tabular}{ c c c } 
 \hline
 Parameter & Value & Description \\
 \hline
 $c_0$ & 2 \microMolar & Total [Ca] in terms of cytosolic volume \\ 
 $c_1$ & 0.185 & Ratio ER volume to cytosol volume \\
 $v_1$ & 6 \text{s}$^{-1}$ & Max Ca channel flux \\
 $v_2$ & 0.11 \text{s}$^{-1}$ & Ca leak flux constant \\
 $v_3$ & 0.9 \microMolarSecInv & Max Ca uptake \\
 $k_3$ & 0.1 \microMolar & Activation constant for ATP-Ca pump \\
 $a_1$ & 400 \microMolarSecInv & $\iptr$ reaction rate \\
 $a_2$ & 0.2 \microMolarSecInv & $\iptr$ reaction rate \\
 $a_3$ & 400 \microMolarSecInv & $\iptr$ reaction rate \\
 $a_4$ & 0.2 \microMolarSecInv & $\iptr$ reaction rate \\
 $a_5$ & 20 \microMolarSecInv & $\iptr$ reaction rate \\
 $d_1$ & 0.13 \microMolar & $\iptr$ reaction rate \\
 $d_2$ & 1.049 \microMolar & $\iptr$ reaction rate \\
 $d_3$ & 943.4 $\times 10^{-3}$ \microMolar & $\iptr$ reaction rate \\
 $d_4$ & 144.5 $\times 10^{-3}$ \microMolar & $\iptr$ reaction rate \\
 $d_5$ & 82.34 $\times 10^{-3}$ \microMolar & $\iptr$ reaction rate \\
 $\mu_0([\cacytM])$ & 0.25 \microMolar & Initial mean Ca concentration \\
 $\mu_0([\ipt])$ & Varying & Initial mean $\ipt$ concentration \\
 $\sigma_0([\cacytM])$ & $10^{-3}$ \microMolar & Initial standard deviation of Ca concentration \\
 $\sigma_0([\ipt])$ & $10^{-3}$ \microMolar & Initial standard deviation of $\ipt$ concentration \\
 $V_\text{Cyt}$ & $10^{-12} \text{ or } 10^{-14}$~L & Cytoplasm volume \\
 $\Delta t^\text{(write)}$ & 0.1 s & Writing interval \\
 $\Delta t^\text{(diff. eq.)}$ & 0.001 s & Integration step length for currents \\
 $T_\text{max}$ & 50 s & Maximum simulation time \\
\end{tabular}
\end{ruledtabular}
\end{table*}


\subsection{Range of oscillations}
\label{app:bifurcations}


Figure~\ref{fig:channels} shows the original bifurcation diagram by~\citet{de_young_1992}, and the range of oscillations in the stochastic model under consideration. 
The stochastic curves should not be interpreted as a bifurcation diagram. Rather, at each concentration of $\ipt$:
\begin{itemize}
\item At each timepoint, average over stochastic simulations to obtain the mean $\mu(t)$ and standard deviation $\sigma(t)$.
\item Calculate the upper and lower curves over time $c_{\pm}(t) = \mu(t) \pm \sigma(t)$.
\item Take the min. and max.:
\begin{equation}
\begin{split}
c_{-,\text{min}} &= \underset{t}{\text{min}}(c_-(t)) , \\
c_{+,\text{max}} &= \underset{t}{\text{max}}(c_+(t)) ,
\end{split}
\end{equation}
over the last $40$s of each of the $100$s simulations.
\end{itemize}
The resulting $c_{-,\text{min}}, c_{+,\text{max}}$ are plotted in Figure~\ref{fig:channels} to indicate the range of oscillations. Error bars indicate $95\%$ confidence levels.


\subsection{Derivation of reaction model from differential equations}
\label{sec:convert}


The original differential equations in~\citet{de_young_1992} are:
\begin{equation}
\frac{d \concCyt{\catpCyt} }{dt} = J_1 - J_2 + J_3
,
\end{equation}
where
\begin{equation}
\begin{split}
J_1 &= c_1 v_2 ( c_1^{-1} \concCyt{\catpER} - \concCyt{\catpCyt} ) 
, \\
J_2 &= \frac{v_3 \concCyt{\catpCyt}^2}{\concCyt{\catpCyt}^2 + k_3^2}
, \\
J_3 &= c_1 v_1 x_{110}^3 ( c_1^{-1} \concCyt{\catpER} - \concCyt{\catpCyt} ) 
,
\end{split}
\end{equation}
where $x_{110}$ is the fraction of subunits in the open state $S_{110}$, and $c_1,v_1,v_2,v_3,k_3$ are constants given in Table~\ref{tab:table1}. The current $J_1$ is a leak current, $J_2$ is the flux out of cytoplasm due to an ATP-dependent $\catp$ pump, e.g. a SERCA pump, and $J_3$ is the transport of $\catp$ into cytoplasm through the open $\iptr$. The notation $[X_\text{ER}]_\text{Cyt}$ is used to denote the number of particles of species $X$ located in the $\text{ER}$, divided by the volume of $\text{Cyt}$ to obtain a concentration (as opposed to the volume of the ER). This conversion is used to simplify the calculations by having to keep track of only a single volume.

The currents $J_1,J_2$ are kept as differential equations, while the transport $J_3$ is converted to an equivalent reaction system. For this transformation, \textit{concentration-based} reaction rates must be transformed into \textit{molecular-based} reaction rates. 

Consider a general reaction of the form: 
\begin{equation}
\sum_{i=1}^R m_i X_i \underset{\gamma}{\rightarrow} \dots
.
\label{eq:rxnBased}
\end{equation}
Associated with this reaction is the stoichiometry vector $\boldsymbol{\nu}$ of length $R$, whose components $\nu_i$ describe the change in the number of particles of $X_i$. Here, $\gamma$ will be referred to as the molecular-based reaction rate. The goal is to relate $\gamma$ to the concentration-based reaction rate $k$ appearing in mass action kinetics:
\begin{equation}
\frac{d [X_j] }{dt} = - \nu_j k \prod_{i=1}^R \left ( [X_i] \right )^{m_i} + \dots
,
\label{eq:massAction}
\end{equation}
where the $\dots$ denote other possible reactions.

The propensity term for the reaction~(\ref{eq:rxnBased}) in units of molecules per time is:
\begin{equation}
\rho = \gamma \prod_{i=1}^R \binom{n_{X_i}}{m_i}
.
\end{equation}
where $n_{X_i}$ is the number of particles of species $X_i$. At large particle numbers, this is commonly approximated by
\begin{equation}
\rho \approx \gamma \prod_{i=1}^R \frac{n_{X_i}^{m_i}}{m_i !}
.
\label{eq:prop}
\end{equation}

In the mass action equation~(\ref{eq:massAction}), the reaction-based rate of change in units of concentration per time is (without the stoichiometry vector):
\begin{equation}
k \prod_{i=1}^R \left ( [X_i] \right )^{m_i}
.
\end{equation}
Substitute the definition of the concentration $[X_i] = n_{X_i} / ( c_A \times V ) $ for $n_{X_i}$ particles in volume $V$ where $c_A$ is Avogadros constant, and convert to units of molecules per time by multiplying by $c_A \times V$:
\begin{equation} 
k c_A V \prod_{i=1}^R \left ( \frac{n_{X_i}}{ c_A V} \right )^{m_i}
.
\end{equation}
Equating this with the approximation for the propensity~(\ref{eq:prop}) gives the relation:
\begin{equation}
\gamma
=
k c_A V
\prod_{i=1}^R \frac{m_i !}{ \left ( c_A V \right )^{m_i} }
.
\label{eq:rateConv}
\end{equation}

Using this relation, the transport current $J_3$ can be transformed by writing it in the equivalent form:
\begin{equation}
\begin{split}
J_3 =& k_{1f} \concCyt{ S_{110} }^3 \concCyt{ \catp_\text{ER} } - k_{1b} \concCyt{ S_{110} }^3 \concCyt{ \catp_\text{Cyt} } , \\
k_{1f} =& v_1 \concCyt{\iptr}^{-3} , \\
k_{1b} =& c_1 v_1 \concCyt{\iptr}^{-3} .
\end{split}
\end{equation}
where $S_{110}$ is the open state of the $\iptr$ in Figure~\ref{fig:channels}. From this the equivalent reaction can be identified:
\begin{equation}
3 S_{110} + \catp_\text{ER} \underset{\gamma_{1b}}{\overset{\gamma_{1f}}{\rightleftharpoons}} 3 S_{110} + \catp_\text{Cyt}
,
\end{equation}
where the molecular-based rates are given by~(\ref{eq:rateConv}):
\begin{equation}
\begin{split}
\gamma_{1f} 
&= 6 v_1 n_\text{IP3R}^{-3} , \\
\gamma_{1b}
&= 6 c_1 v_1 n_\text{IP3R}^{-3} ,
\end{split}
\end{equation}
where $n_\text{IP3R}$ is the number of $\iptr$.


\subsection{Number of IP3 receptor subunits}


The number of $\ipt$ receptors ($\iptr$) depends on the density of channels and the surface area of the ER. The ER is a highly folded structure, and as such its surface area can vary significantly. In the text a large spread in the number of channels is explored. Here, an order of magnitude estimation is provided to justify their scale.

\begin{figure*}[htp]
	\centering
	\includegraphics[width=1.0\textwidth]{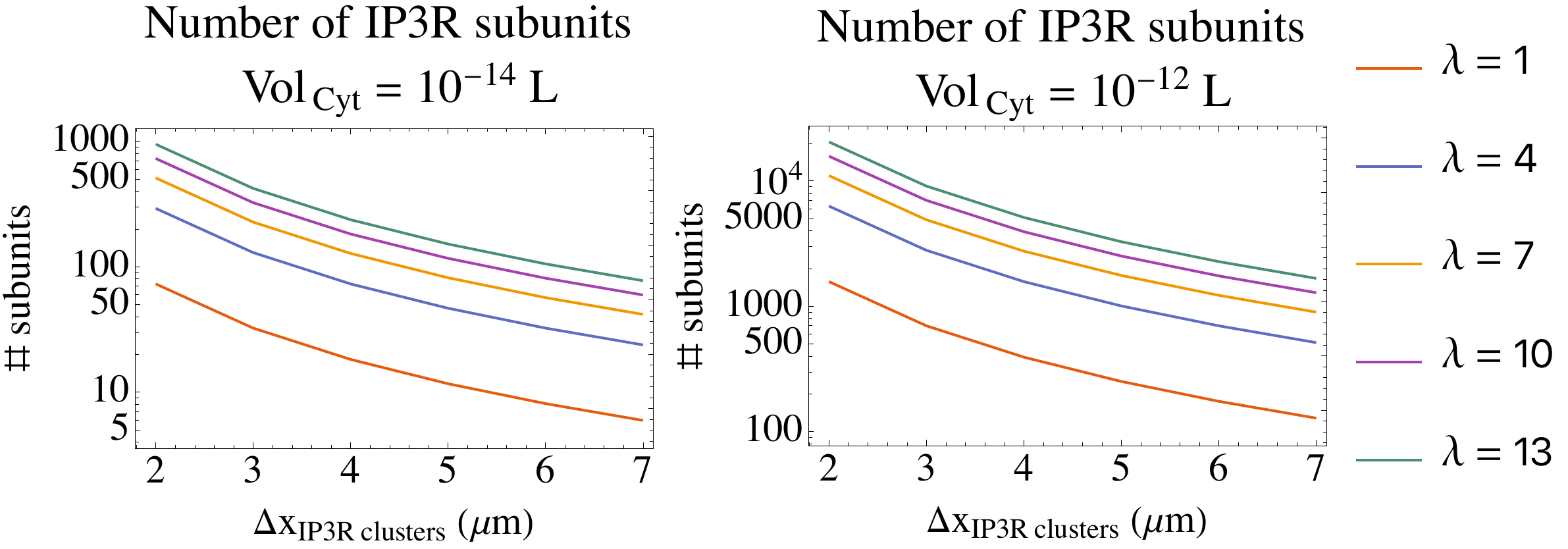}
	\caption{Number of IP3R subunits given by~(\ref{eq:noIP3R}) as a function of the cluster spacing $\Delta x_\text{IP3R clusters}$ in $\mu \text{m}$, and the dimensionless surface area factor $\lambda$, where $\lambda = 1$ corresponds to a sphere.}
	\label{fig:noIP3R}
\end{figure*}

Starting with the volume of the cytoplasm $V_\text{cyt}$, the volume of the ER is $V_\text{ER} = c_1 \times V_\text{cyt}$ where $c_1 = 0.185$ is the ratio estimated in~\citet{de_young_1992}. The ER has the smallest surface area if it is a sphere:
\begin{equation}
\text{SA}_\text{ER}^\text{min} = 4 \pi \left ( \frac{3}{4\pi} V_\text{ER} \right )^{2/3}
.
\end{equation}
Let the actual surface area be some factor $\lambda$ larger than the minimum:
\begin{equation}
\text{SA}_\text{ER} = \lambda \times \text{SA}_\text{ER}^\text{min}
.
\end{equation}
$\iptr$ clusters are spread out over the ER with spacing $1-7\mu\text{m}$~\cite{voorsluijs_2019}. Assuming the IP3R clusters were are located in a grid with spacing $\Delta x_\text{IP3R}$ gives:
\begin{equation}
n_\text{IP3R clusters} = \frac{\text{SA}_\text{ER}}{\Delta x_\text{IP3R clusters}^2}
.
\end{equation}
Each cluster contains up to 15 channels~\cite{voorsluijs_2019}. Assuming 10 channels per cluster and with 4 subunits per channel gives:
\begin{equation}
n_\text{IP3R subunits} 
= 10 \times 4 \times n_\text{IP3R clusters} 
= \frac{ 160 \pi \lambda }{ \Delta x_\text{IP3R clusters}^2 } \left ( \frac{3}{4 \pi} c_1 V_\text{cyt} \right )^{2/3}
.
\label{eq:noIP3R}
\end{equation}
Figure~\ref{fig:noIP3R} shows the number of $\iptr$ subunits for a range of spacings and surface area factors. For a highly folded ER with large $\lambda$ and average cluster spacing $\Delta x_\text{IP3R clusters} \sim 3 \mu\text{m}$, the estimates of $\mathcal{O}(100)$ subunits for $\text{Vol}_\text{cyt} = 10^{-14} L$ and $\mathcal{O}(1000)$ subunits for $\text{Vol}_\text{cyt} = 10^{-12} L$ are reasonable, which are the approximate magnitudes explored in the text.

\clearpage
\section{Training ML models} \label{app:training}



\subsection{Data transformation}


\begin{figure}[htp]
	\centering
	\includegraphics[width=0.7\textwidth]{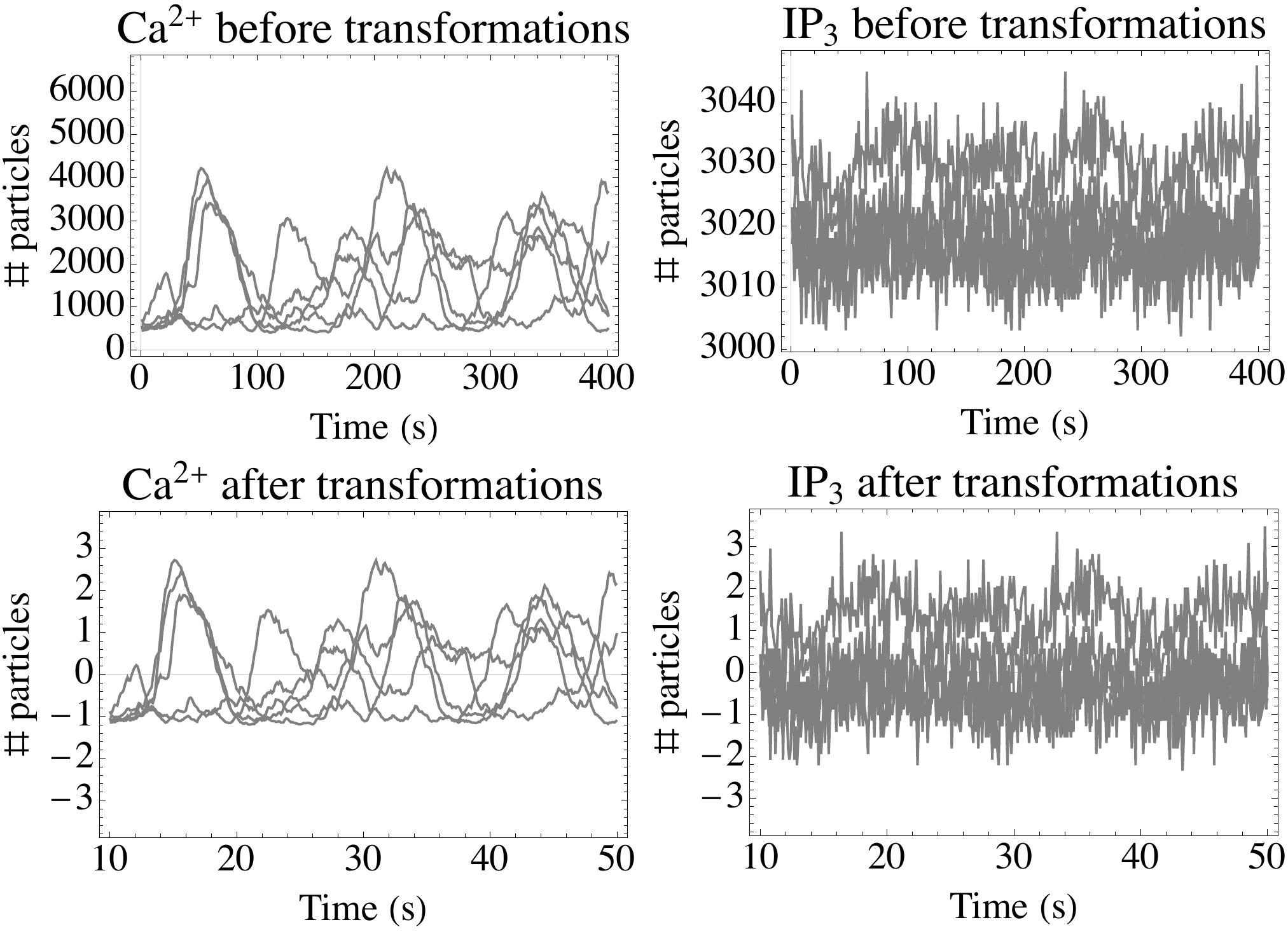}
	\caption{Example standardizing transformation for the system studied in \figThree at $[\ipt] = 0.5 \mu$M. Five stochastic simulation trajectories are shown in gray. \textit{Top row}: Number of particles of calcium and $\ipt$ before transformation. \textit{Bottom row}: After transformation, oscillations occur around zero.}
	\label{fig:ex_trans}
\end{figure}

\begin{figure}[htp]
	\centering
	\includegraphics[width=1.0\textwidth]{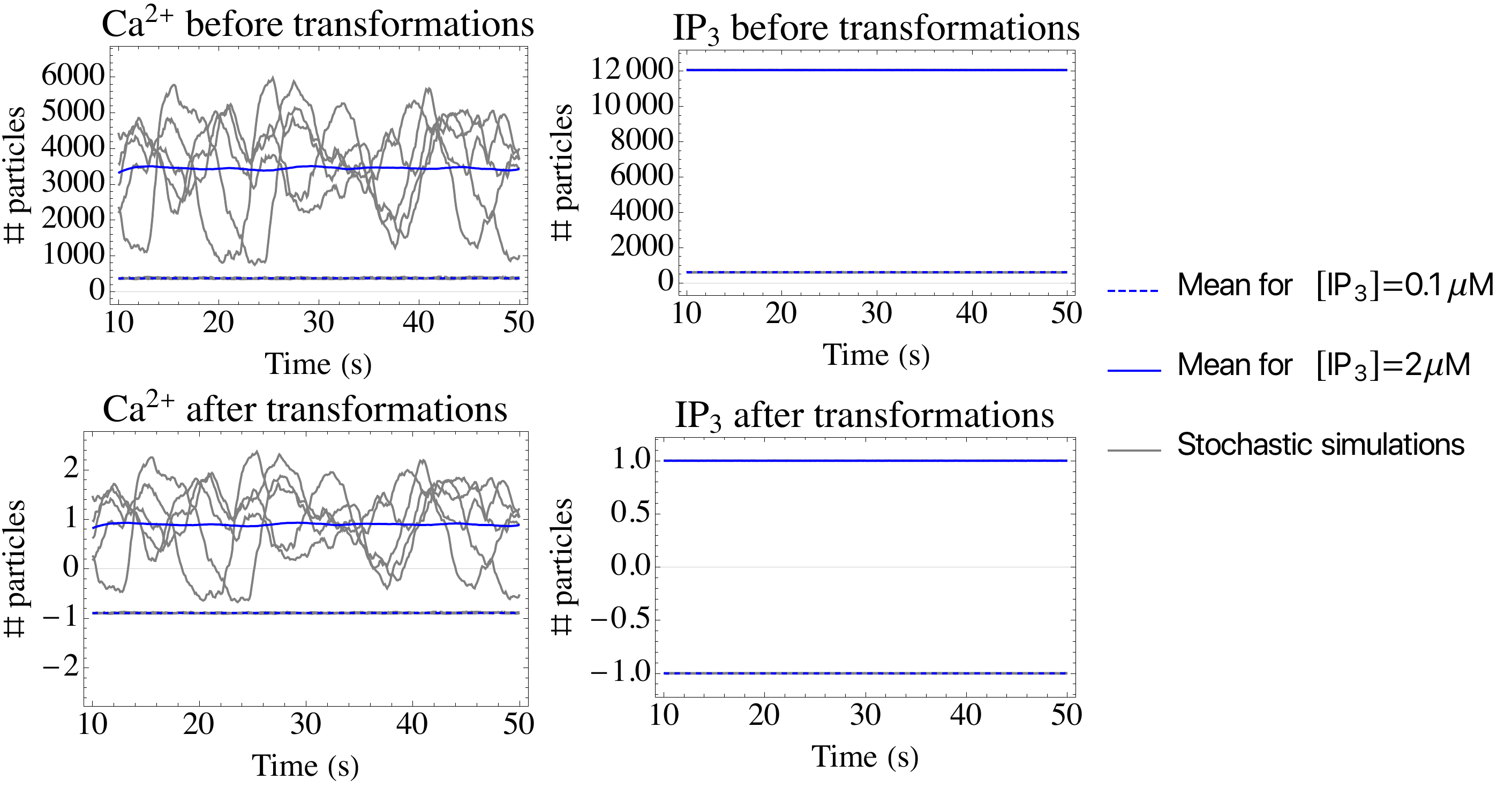}
	\caption{
	Transformations for the $\ipt$ system studied in \figThree.
	The transformation is calculated for stochastic simulations at the upper and lower range of oscillations considered, i.e. $[\ipt] = 0.1 \mu$M and $[\ipt] = 2 \mu$M.
	}
	\label{fig:trans}
\end{figure}

\begin{figure*}[htp]
	\centering
	\includegraphics[width=1.0\textwidth]{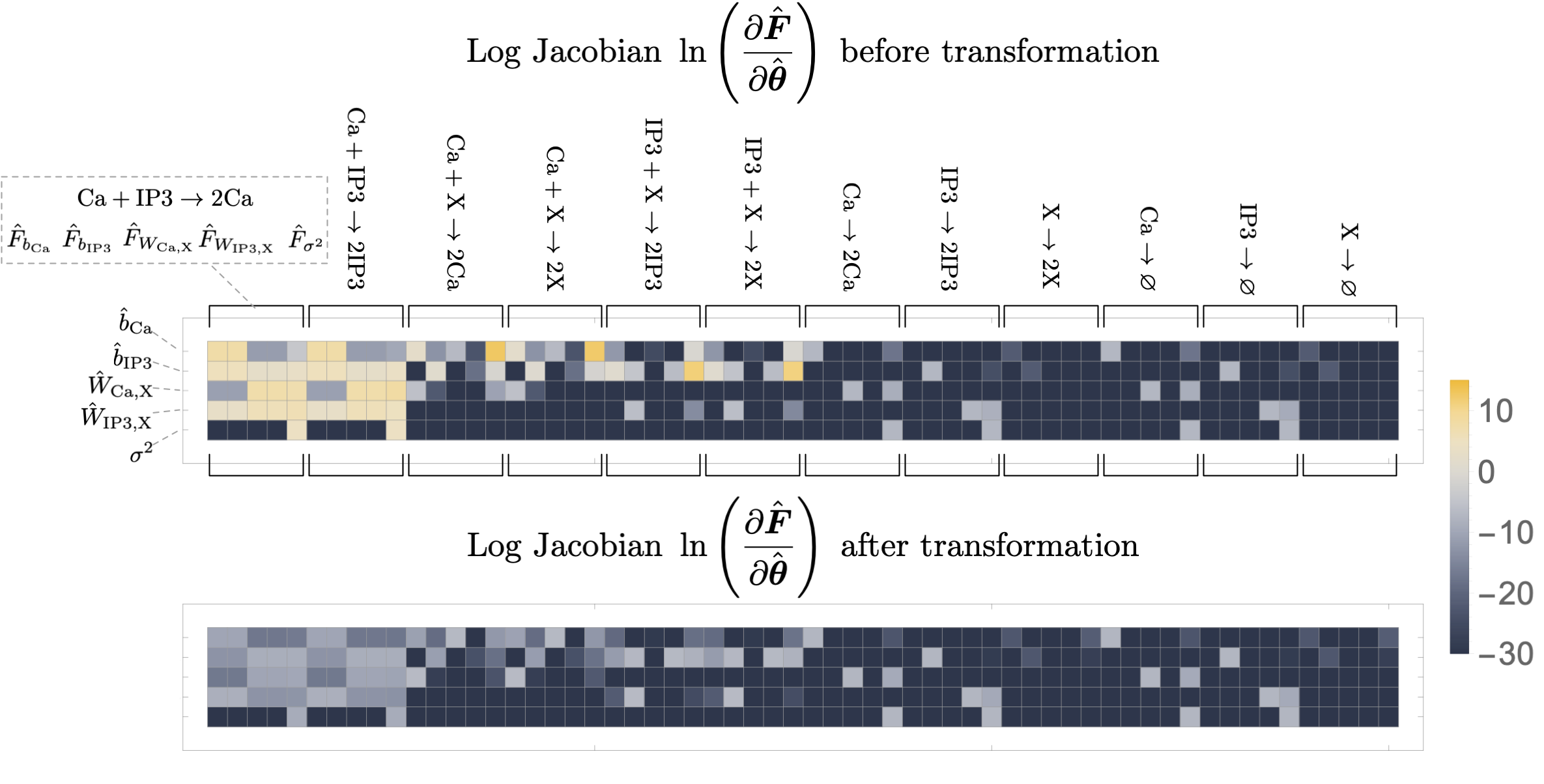}
	\caption{Jacobian of reaction candidates with respect to standard parameters, plotted on a log scale. The reactions are those of the model studied in \figThree, and parameters $\hat{\thetabf}$ are those at a single point in time $t=50$ and $[\ipt] = 0.7 \mu$M, with $\mu_h = 0$ and $\Sigma_h = I$. \textit{Top:} Jacobian before the transformation~(\ref{eq:transRes}). \textit{Bottom:} Jacobian after the transformation. The transformation reduces the Jacobian, increasing the stability of the integration because small perturbations do not significantly alter the inputs to the neural network.}
	\label{fig:jac}
\end{figure*}

Let the stochastic simulation data be represented by the matrix $X(t)$ of size $M \times N_v$ where $N_v$ is the dimension of the visible variables and $M$ is the number of samples:
\begin{equation}
X(t) = \begin{pmatrix}
	\boldsymbol{x}_1^\intercal(t) \\
	\boldsymbol{x}_2^\intercal(t) \\
	\dots \\
	\boldsymbol{x}_M^\intercal(t)	
\end{pmatrix}
.
\end{equation}
PCA applied to $X(t)$ leads to the parameters:
\begin{equation}
\hat{\thetabf}_X(t) = \{ \hat{\bbf}_X, \hat{W}_X, \sigma_X^2 \}(t)
.
\end{equation}
From these parameters, approximations $\hat{\Fbf}_\text{rxn.}^\text{(approx.)}$ to $\hat{\Fbf}$ under different reaction processes can be calculated.

When using the model to integrate the parameters $\hat{\thetabf}_X(t)$, it traces out a trajectory in $\hat{D} = N_v + N_v \times N_h + 1$ dimensional space. In order for this integration to be stable, the inputs to the neural network $\hat{\Fbf}_\text{rxn.}^\text{(approx.)}$ must not be sensitive to small perturbations in $\hat{\thetabf}_X(t)$. Note that the error in $\hat{\thetabf}_X(t)$ is set by the error in the output of the neural network, and the integration drift that arises from integrating a noisy signal.

To ensure that the Jacobian $\partial \hat{\Fbf}_\text{rxn.}^\text{(approx.)} / \partial \hat{\thetabf}_X(t)$ is small, introduce the following transformation. For a given trajectory $\boldsymbol{x}_i(t), i=1,\dots,M$, compute the mean and variance over time and samples:
\begin{equation}
\begin{split}
\mbf
&= \frac{1}{T} \sum_{t=1}^T \frac{1}{M} \sum_{i=1}^M \boldsymbol{x}_i(t) 
, \\
\vbf
&= \frac{1}{T} \sum_{t=1}^T \frac{1}{M} \sum_{i=1}^M [ \boldsymbol{x}_i(t) - \mbf ]^2
,
\end{split}
\label{eq:transDefns}
\end{equation}
and use these parameters to define a transformation:
\begin{equation}
\boldsymbol{y}_i(t) = \frac{ \boldsymbol{x}_i(t) - \mbf }{ \sqrt{ \vbf } }
.
\label{eq:transRes}
\end{equation}
This leads to a new matrix $Y(t)$ of equal size $M \times N_v$. PCA applied to $Y(t)$ leads to a different set of parameters:
\begin{equation}
\hat{\thetabf}_Y(t) = \{ \hat{\bbf}_Y, \hat{W}_Y, \sigma_Y^2 \} (t)
.
\end{equation}

Figure~\ref{fig:ex_trans} shows how such a transformation standardizes oscillations. Figure~\ref{fig:trans} shows the transformation for the system studied in \figThree. Here the averages in~(\ref{eq:transDefns}) are taken over $\ipt$ concentrations at the boundaries of the bifurcation diagram studied, i.e. at $[\ipt] = 0.1 \mu$M and $[\ipt] = 2 \mu$M. 

It is difficult to relate $\thetabf_Y(t)$ to $\thetabf_X(t)$ since it relates the eigendecomposition of a matrix product. Importantly, however, Figure~\ref{fig:jac} shows that the Jacobian $\partial \hat{\Fbf}_\text{rxn.}^\text{(approx.)} / \partial \hat{\thetabf}(t)$ has decreased for the example problem studied in \figThree.

\subsection{Training inputs and targets}


After transforming the data $X \rightarrow Y$, the ML parameters are identified from $Y(t)$ for $t=1,\dots,T$ using \eqML. For the models considered in this work in \figThree and \figFour, the data used are in the range $10$s to $50$s of the stochastic simulations, which are of length $50$s. The first $10$s are discarded to lessen the dependence of the oscillations on the chosen initial condition in Table~\ref{tab:table1}. The parameters obtained are the standard parameters $\hat{\thetabf}(t)$. Note that the sign of the eigenvectors is adjusted to be consistent by ensuring that for any eigenvector $\ubf$ we have $| \cos^{-1} ( \ubf^\intercal \boldsymbol{1} ) | \leq \pi / 2$.

The targets derivatives are calculated from~$\hat{\thetabf}(t)$ using total variation regularization (TVR)~\cite{xiao_2011}. For a time series $\zbf$ with elements $z_1, z_2, \dots, z_T$, the time derivative $\dot{\zbf}$ is obtained by solving the optimization problem:
\begin{equation}
\dot{\zbf} = \underset{\ubf}{\text{min}} \left ( \alpha || \dot{\ubf} ||_1 + \frac{1}{2} || A \ubf - \zbf ||_2 \right ) ,
\end{equation}
where $A$ is the anti-differentiation matrix, and $\alpha$ is a regularization parameter. In this case, the $A$ matrix is that of the Euler method. The optimization problem is solved using the lagged diffusivity method~\cite{xiao_2011} for $10$ optimization steps for every parameter $\zbf$ in $\hat{\thetabf}$ with regularization parameter $\alpha = 100$. Additionally, after calculating the derivatives $d \hat{\thetabf} / dt$, small derivative values with an absolute value below $10^{-5}$ are set to zero. The target outputs $d \hat{\thetabf} / dt$ are therefore obtained:
\begin{equation}
\text{Target}(t) 
= \frac{d \hat{\thetabf}(t) }{ dt }
= \begin{pmatrix}
d \hat{\bbf} / dt \\
d \hat{W} / dt \\
d \sigma^2 / dt
\end{pmatrix}
.
\end{equation}

The inputs are obtained by integrating $d \hat{\thetabf} / dt$ with the anti-differentiation matrix $A$ to obtain smoothed inputs $\hat{\thetabf}^\text{(integrated)}$:
\begin{equation}
\text{Input}(t)
= 
\hat{\thetabf}^\text{(integrated)} (t)
=
\begin{pmatrix}
\hat{\bbf} (t) \\
\hat{W} (t) \\
\sigma^2 (t)
\end{pmatrix}
.
\end{equation}


\subsection{Reaction approximations}


\begin{figure}[htp]
	\centering
	\includegraphics[width=0.75\textwidth]{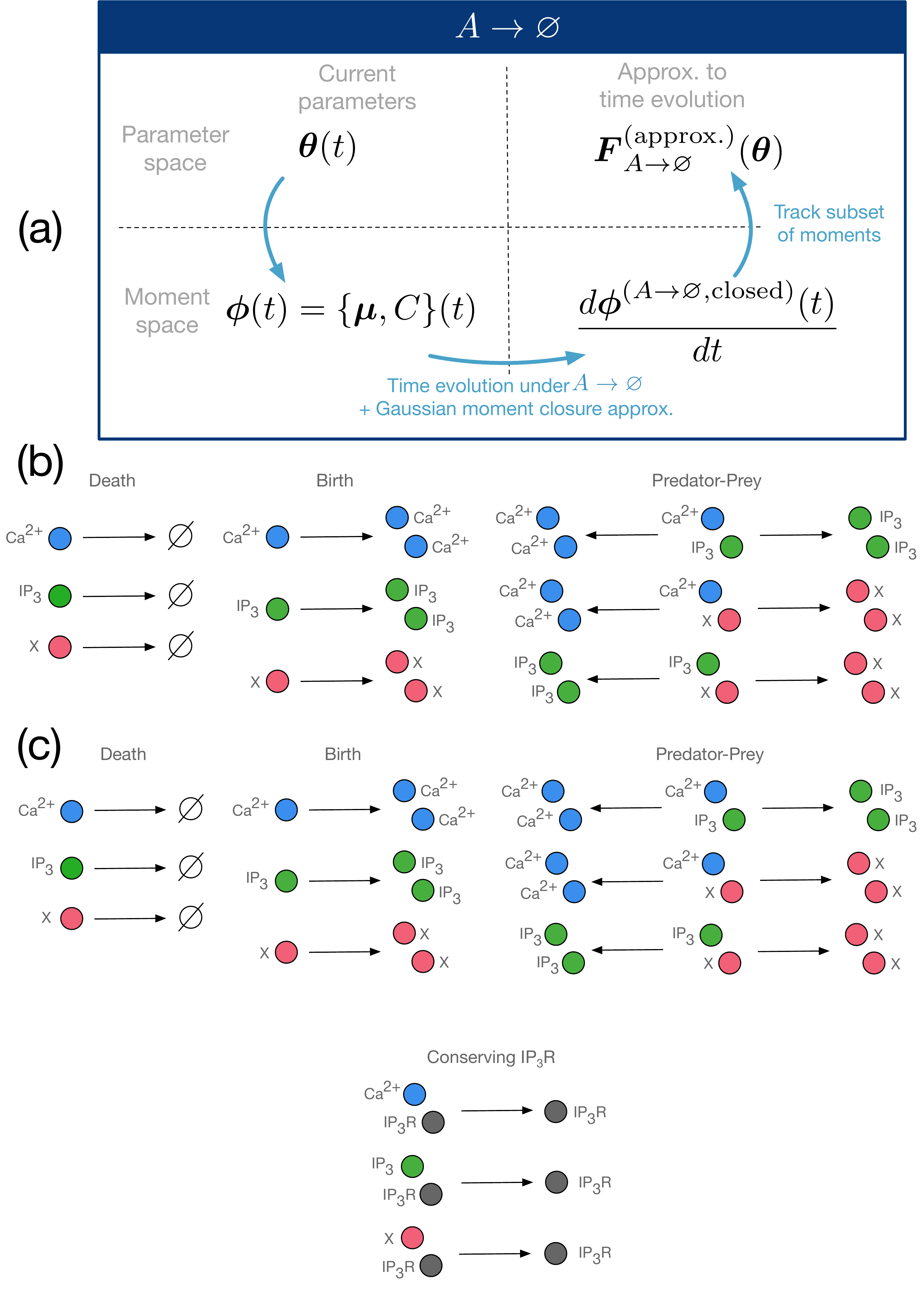}
	\caption{
	(a) Schematic of how an approximation to the time evolution of parameters is derived for a single reaction pathway. The observables $\phibf(t)$ corresponding to the distribution defined by the current parameters $\thetabf(t)$ at an instant in time are obtained from \eqPCA.
	The time evolution of each observable is calculated from the master equation, and closed using a Gaussian moment closure approximation (\eqGaussClosure).
	By tracking exactly only a subset of moments $d \{ \mubf_v, \Sigma_{vh}, \text{Tr}(\Sigma_v), \mubf_h, \Sigma_h \} / dt$ matching the dimensions of $\thetabf(t) = \{ \bbf, W, \sigma^2, \mubf_h, \Sigma_h \}(t)$, an approximation to the time evolution of the parameters is obtained.
	(b) The reactions used to construct the inputs to the sub-network models in \figThree generalizing in $\ipt$: death e.g. $H \rightarrow \varnothing$, birth e.g. $P \rightarrow 2P$, and a predator-prey interaction, e.g. $H + P \rightarrow 2 H$. These reactions mimic the reaction scheme of Lotka-Volterra system. Each combination of $\{ H, P \}$ from $\{ \catp, \ipt, X \}$ is included.
	(c) The reactions used in the models including $\iptr$. The reactions are those of (b), including three extra reactions for $\iptr$ that explicitly conserve $\iptr$.
	}
	\label{fig:rxnInputs}
\end{figure}

The physics of the system is described by the chemical master equation (CME). The CME can be incorporated into the inference problem by using it to derive an approximation to the time evolution of parameters. Figure~\ref{fig:rxnInputs} shows a schematic of how this derivation as follows.

For the distribution defined by the current parameters $\thetabf(t) = \{ \bbf, W, \sigma^2, \mubf_h, \Sigma_h \}(t)$ at an instant in time, \eqPCA gives the observables $\phibf(t) = \{ \mubf, C \}(t)$. Next, consider for example the predator-prey reaction $H + P \rightarrow 2H$ with rate $k$ for predators $H$ and prey $P$. Under this reaction, the observables evolve in time according to:
\begin{equation}
\begin{split}
\frac{d \left \langle n_P \right \rangle}{dt} &= - k \langle n_H n_P \rangle , \\
\frac{d \left \langle n_H \right \rangle}{dt} &= k \langle n_H n_P \rangle ,
\end{split}
\end{equation}
and
\begin{equation}
\begin{split}
\frac{d \left \langle n_P^2 \right \rangle}{dt} &= - 2 k \langle n_H n_P^2 \rangle + k \langle n_H n_P \rangle , \\
\frac{d \left \langle n_H^2 \right \rangle}{dt} &= 2 k \langle n_H^2 n_P \rangle + k \langle n_H n_P \rangle , \\
\frac{d \left \langle n_H n_P \right \rangle}{dt} &= - k \langle n_H^2 n_P \rangle + k \langle n_H n_P^2 \rangle - k \langle n_H n_P \rangle ,
\end{split}
\end{equation}
and for $X \notin \{ H, P \}$
\begin{equation}
\begin{split}
\frac{d \left \langle n_X n_P \right \rangle}{dt} &= - k \langle n_X n_H n_P \rangle , \\
\frac{d \left \langle n_X n_H \right \rangle}{dt} &= k \langle n_X n_H n_P \rangle ,
\end{split}
\end{equation}
which are derived from the CME. These equations form the time evolution of the observables under this reaction $d \phibf^{(H+P\rightarrow 2H)} / dt$. In the derivation of the desired approximations, the reaction rates are set to unity $k=1$. Ultimately, if a linear model is learned instead of a neural network, the learned coefficients can be interpreted as the learned reaction rates.

These equations are not closed - higher order observables appear on the right hand side. In principle, any moment closure approximation can be used to derive an approximation, but the natural choice is to use Gaussian moment closure (\eqGaussClosure) since the model is Gaussian. The space of candidates can also be increased by deriving approximations for more than one closure approximation. Under this approximation, for any species $X,Y,Z$:
\begin{equation}
\begin{split}
\langle n_X n_Y n_Z \rangle 
\rightarrow &
- 2 \langle n_X \rangle \langle n_Y \rangle \langle n_Z \rangle
+ \langle n_X \rangle \langle n_Y n_Z \rangle
+ \langle n_Y \rangle \langle n_X n_Z \rangle
+ \langle n_Z \rangle \langle n_X n_Y \rangle .
\end{split}
\end{equation}
This approximation gives the closed form for the observables $d \phibf^{(H+P\rightarrow 2H, \text{closed})} / dt$.

To convert back to the parameter frame, we note that the transformation $d \phibf / dt \rightarrow d \thetabf / dt$ may not exist. Instead, we track only a certain number of observables equivalent to the number of parameters $D$ in the model. While the choice of the observables is arbitrary, the obvious choice is those that match the dimensions of the parameters:
\begin{equation}
\frac{d}{dt} \left \{ \mubf_v, C_{vh}, \text{Tr}(C_v), \mubf_h, \Sigma_h \right \} ,
\end{equation}
which match the dimensions of $\thetabf(t) = \{ \bbf, W, \sigma^2, \mubf_h, \Sigma_h \}(t)$. The transformations are obtained by differentiating \eqTransHat:
\begin{equation}
\begin{split}
\frac{d W^\intercal}{dt} = & - \Sigma_h^{-1} \frac{d \Sigma_h}{dt} \Sigma_h^{-1} C_{vh} + \Sigma_h^{-1} \frac{d C_{vh}}{dt} , \\
\frac{d \bbf}{dt} = & \frac{d \mubf_v}{dt} - \frac{d C_{vh}^\intercal}{dt} \Sigma_h^{-1} \mubf_h
+ C_{vh}^\intercal \Sigma_h^{-1} \frac{d \Sigma_h}{dt} \Sigma_h^{-1} \mubf_h
- C_{vh}^\intercal \Sigma_h^{-1} \frac{d \mubf_h}{dt} , \\
\frac{d \sigma^2}{dt}
= & \frac{d \text{Tr}(C_v)}{dt} - \text{Tr} \Bigg (
\left ( \frac{d C_{vh}}{dt} \right )^\intercal \Sigma_h^{-1} \Sigma_{vh} + C_{vh}^\intercal \Sigma_h^{-1} \frac{d C_{vh}}{dt}
- C_{vh}^\intercal \Sigma_h^{-1} \frac{d \Sigma_h}{dt} \Sigma_h^{-1} C_{vh} \Bigg ) .
\end{split}
\end{equation}
The resulting time evolution vector $\Fbf_{H+P\rightarrow 2H}^\text{(approx.)}  (\thetabf(t))$ is an approximation to the true time evolution. It has been shown that the linearity of the CME in reaction operators extends this form of the approximation~\cite{ernst_2018}. If a linear combination of such approximations is used instead of a neural network, the learned coefficients are directly the reaction rates associated with each process.

Finally, the differential equations can be transformed to the standard parameter space using the inverse of \eqTransHat and its derivative:
\begin{equation}
\begin{split}
\hat{\bbf} &= \bbf + W \mubf_h , \\
\hat{W} &= W \Sigma_h^{1/2} ,
\end{split}
\end{equation}
and
\begin{equation}
\begin{split}
\hat{F}_{\hat{\bbf}} &= F_{\bbf} + F_W \mubf_h + W F_{\mubf_h} , \\
\hat{F}_{\hat{W}} &= F_W \Sigma_h^{1/2} + \frac{1}{2} W \Sigma_h^{-1/2} F_{\Sigma_h},
\end{split}
\label{eq:transDeriv}
\end{equation}
where~(\ref{eq:transDeriv}) only holds for diagonal latent covariance matrices $\Sigma_h$ as parameterized in \eqFourier due to the derivative of the matrix square root. In general, the derivative of the matrix square root may be expressed through a Kronecker sum as $d \Sigma_h^{1/2} / dt = (\Sigma_h^{1/2} \oplus \Sigma_h^{1/2})^{-1}$.

The result of the transformation is $\hat{\Fbf}_{H+P\rightarrow 2H}^\text{(approx.)}$, the approximation to the time evolution in the standard space.


\subsection{Size of computer algebra expressions}

\begin{figure}[!ht]
	\centering
	\includegraphics[width=1.0\textwidth]{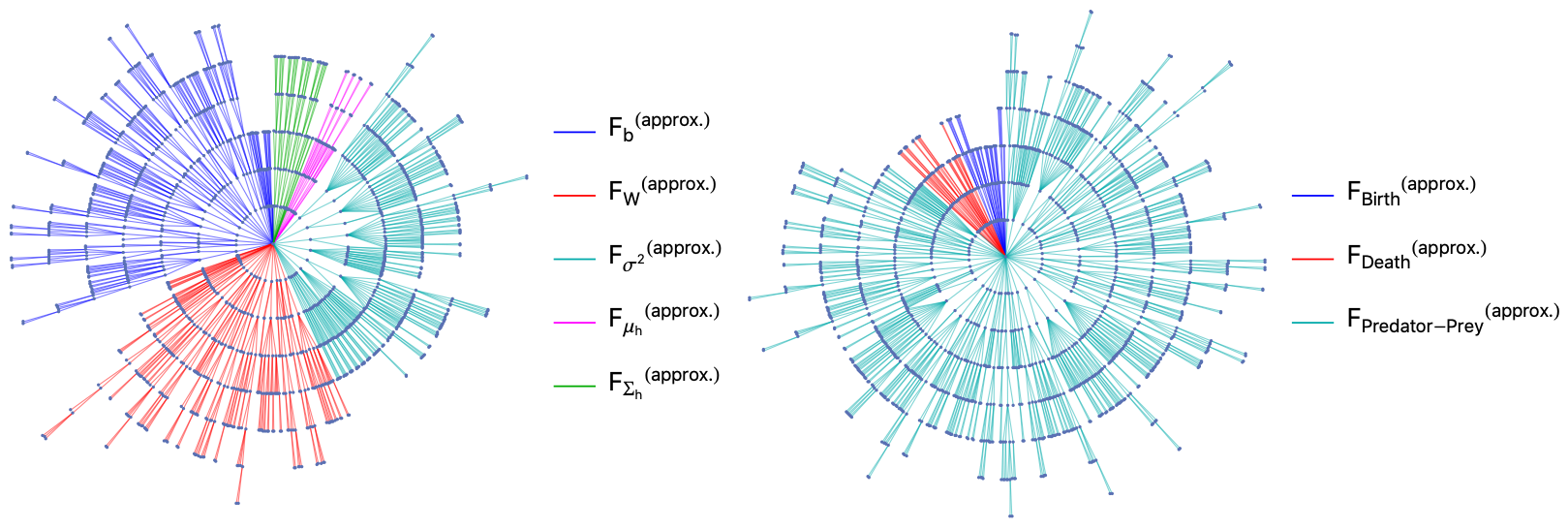}
	\caption{Visualization of the syntax tree for the reaction approximations of Figure~\ref{fig:rxnInputs}(b). Each edge originating from the center node is an individual reaction approximation. \textit{Left:} arranged by PCA parameter $\thetabf$. \textit{Right:} arranged by reaction type.}
	\label{fig:syntaxTree}
\end{figure}

To calculate the reaction approximations, a computer algebra system (Mathematica) was used to derive the expressions. To estimate the size of the analytic expressions, we calculate the number of nodes in the abstract syntax tree for the different reactions considered in Figure~\ref{fig:rxnInputs}(b), after using Mathematica to simplify the expressions. The total number of nodes across all reactions for the calculation $\thetabf \rightarrow \Fbf_\text{reaction}^\text{(approx.)}$ is $1442$ nodes. The average number of nodes per reaction is $120$. Figure~\ref{fig:syntaxTree} visualizes the syntax tree for all reactions by reaction and by PCA parameter. This measure of the 
functional forms $ \Fbf$ quantifies the amount of 
domain-specific knowledge
brought into the method by 
the reaction-based model,
beyond what's in 
the parameter-based model.

\subsection{Standardizing inputs / outputs of subnet}


The inputs to the subnet of \figOne are standardized as is typical for neural networks, as well as the target outputs. From the training data $X(t)$ of size $M$ samples by $N_v$ visible variables, the matrix is transformed $X(t) \rightarrow Y(t)$ as discussed previously, and then the standard ML parameters $\hat{\thetabf}_\text{ML} (t)$ are obtained. 

The standardization for the outputs is straightforward. By differentiating $\hat{\thetabf} (t)$ with total variation regularization, the target outputs of the subnet $d \hat{\thetabf}_\text{ML} / dt$ are obtained. The mean and standard deviation are calculated:
\begin{equation}
\begin{split}
\mubf^\text{(targets)} &= \frac{1}{T} \sum_{t=1}^T \frac{d \hat{\thetabf}_\text{ML} (t) }{dt} , \\
\boldsymbol{\sigma}^\text{(targets)} &= \left ( \frac{1}{T} \sum_{t=1}^T \left ( \frac{d \hat{\thetabf}_\text{ML} (t) }{dt} - \mubf^\text{(targets)} \right )^2 \right )^{1/2} ,
\end{split}
\end{equation}
and the targets are standardized:
\begin{equation}
\frac{d \hat{\thetabf}_\text{ML} (t) }{dt} \rightarrow \left ( \frac{d \hat{\thetabf}_\text{ML} (t) }{dt} - \mubf^\text{(targets)} \right ) / \boldsymbol{\sigma}^\text{(targets)} .
\end{equation}

Standardizing the inputs is not straightforward because they depend on the latent parameters $\mubf_h,\Sigma_h$, which are defined by their Fourier coefficients~(\eqFourier), which are learned. The standardizing parameters may be learned with a normalizing layer, but this is challenging in practice. Instead, we bootstrap the inputs to estimate these parameters as follows. Keep only the highest frequency $f_\text{max.}$ in the Fourier expansion~(\eqFourier), and set all corresponding coefficients $a = b = 1$. Then the ML parameters are bootstrapped with this choice for $\mubf_h,\Sigma_h$ to estimate the inputs $\hat{\Fbf}_{A+B\rightarrow 2B}^\text{(approx.)}$ for the different reactions, and the standardization proceeds for each reaction in the usual way:
\begin{equation}
\begin{split}
\mubf_{A+B\rightarrow 2B}^\text{(inputs)} &= \frac{1}{T} \sum_{t=1}^T \hat{\Fbf}_{A+B\rightarrow 2B}^\text{(approx.)} (\hat{\thetabf}(t)) , \\
\boldsymbol{\sigma}_{A+B\rightarrow 2B}^\text{(inputs)} &= \left ( \frac{1}{T} \sum_{t=1}^T \left ( \hat{\Fbf}_{A+B\rightarrow 2B}^\text{(approx.)} (\hat{\thetabf}(t)) - \mubf_{A+B\rightarrow 2B}^\text{(inputs)} \right )^2 \right )^{1/2} , \\
\hat{\Fbf}_{A+B\rightarrow 2B}^\text{(approx.)} &\rightarrow \left ( \hat{\Fbf}_{A+B\rightarrow 2B}^\text{(approx.)} - \mubf_{A+B\rightarrow 2B}^\text{(inputs)} \right ) / \boldsymbol{\sigma}_{A+B\rightarrow 2B}^\text{(inputs)} .
\end{split}
\end{equation}


\clearpage

\section{Learned model of Calcium oscillations} \label{app:trained}



\subsection{Frequencies}


The latent mean and variance are learned in parallel to the parameters in the differential equation model. By not fixing these parameters, the model is more easily able to find a non-intersecting trajectory in $\thetabf$-space. The latent parameters $\mu_h,\Sigma_h$ are represented by the Fourier decomposition given in \eqFourier.

\begin{figure}[htp]
	\centering
	\includegraphics[width=1.0\textwidth]{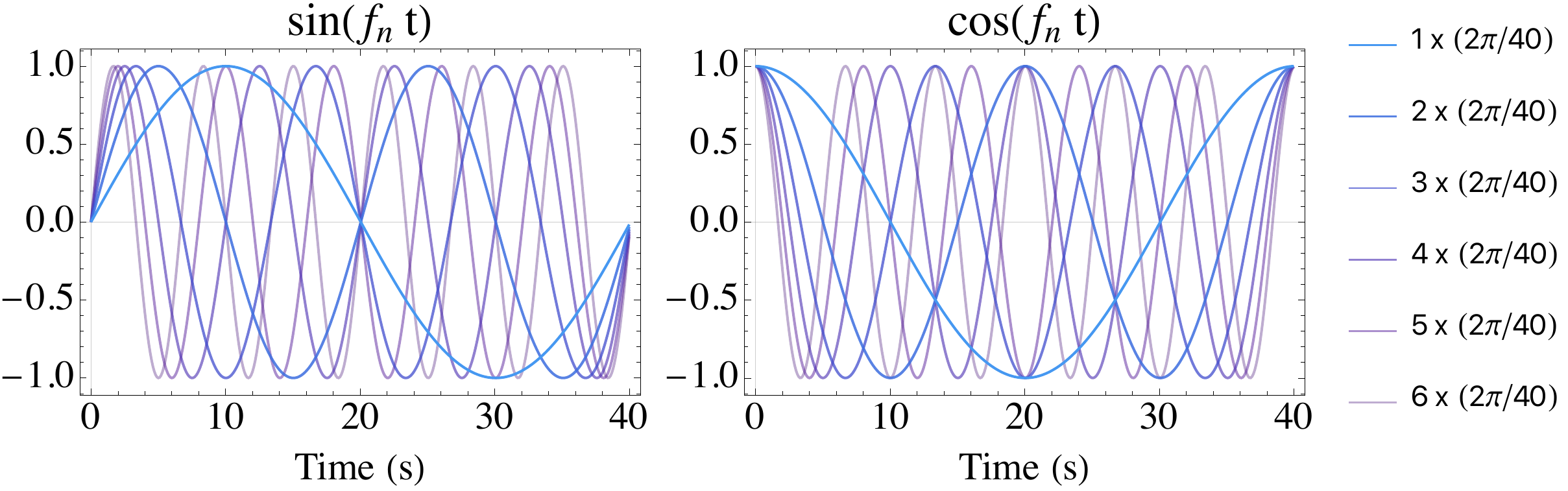}
	\caption{Frequencies in \eqFourier used to train the DBD models. The frequencies are chosen to represent oscillations on the same order of magnitude as the calcium oscillations over several seconds.}
	\label{fig:freqs}
\end{figure}

Figure~\ref{fig:freqs} shows the frequencies chosen for calcium oscillation models (\figThree and \figFour). The frequencies $f_n = \{ 1,2,3,4,5,6 \} \times 2\pi / 40$ $s^{-1}$ allow oscillations on the same period as the calcium oscillations over several seconds.


\subsection{Learned latent representation}

\begin{figure}[htp]
	\centering
	\includegraphics[width=0.6\textwidth]{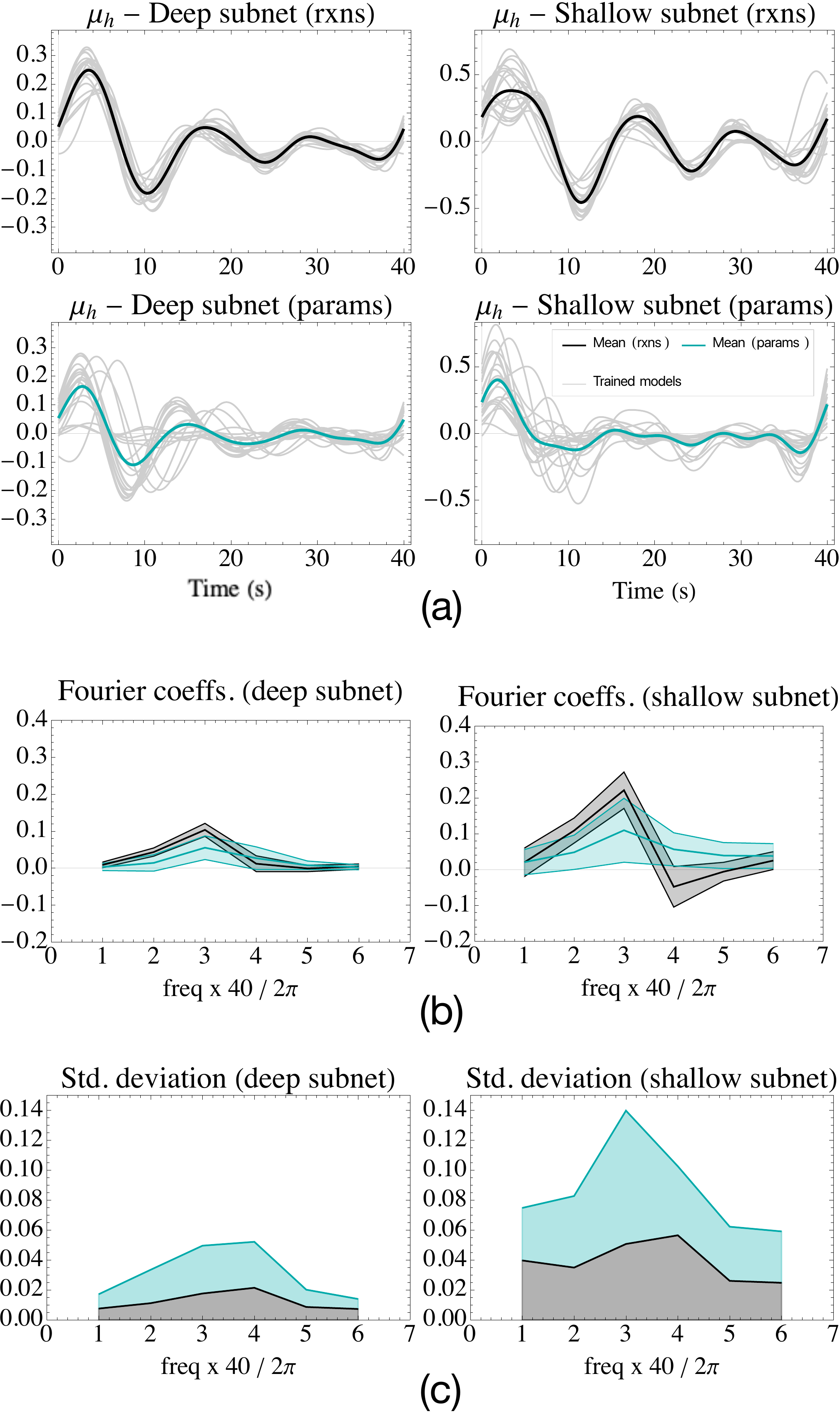}
	\caption{
	(a) Learned $\mu_h$ time evolution according to the Fourier representation~(\eqFourier) for $20$ identical trained models and their mean. \textit{Left column:} deep \& wide subnet as shown in \figThree; \textit{bottom row:} shallow \& thin subnet. \textit{Right column:} reactions based model as shown in \figOne; \textit{right column:} comparison model without reaction approximations as shown in Figure~\ref{fig:compare}. 
	(b) The learned Fourier coefficients for $\mu_h$.
	(c) The standard deviation of the coefficients. The coefficients in the reaction based model are more coherent.
	}
	\label{fig:fourier}
\end{figure}

Figure~\ref{fig:fourier} shows the learned latent mean $\mu_h$ in this representation for the four models of \figThree: a deep \& wide subnet compared to a shallow \& thin subnet, with each a reaction-based model compared to parameter-only model without reactions. The learned frequencies for the reaction-based model are more coherent than for the parameter-only model as seen in Figure~\ref{fig:fourier}, and similarly for $\Sigma_h$. This coherence suggests that the network uncovers an emergent order parameter.


\subsection{Learned moment closure approximation}


DBDs learn a moment closure approximation from data. The reduced model evolves in time as:
\begin{equation}
\begin{split}
\frac{d \pt}{dt} = \sum_{\theta \in \thetabf} \frac{\partial \pt}{\partial \theta} \times F_\theta ,
\end{split}
\end{equation}
where $F_\theta (\thetabf) = d \theta / dt$ by definition. An observable $\langle X(\nbf) \rangle$ where $X(\nbf)$ is some scalar function evolves according to:
\begin{equation}
\begin{split}
\frac{d \langle X(\nbf) \rangle}{dt} = \sum_{\theta \in \thetabf} \left ( \sum_{\nbf} X(\nbf) \frac{\partial \pt}{\partial \theta} \right ) \times F_\theta .
\end{split}
\end{equation}
For maximum-entropy distributions, this quantity is a covariance between $X(\nbf)$ and the moments controlled by interactions in the energy function~\cite{ernst_2019_arxiv}. For example, for a restricted Boltzmann machine, correlations between visible units have been replaced by correlations with latent variables, whose activation is learned. For the Gaussian distribution of PCA, it is easiest to work out the equations numerically.

\begin{figure*}[htp]
	\centering
	\includegraphics[width=1.0\textwidth]{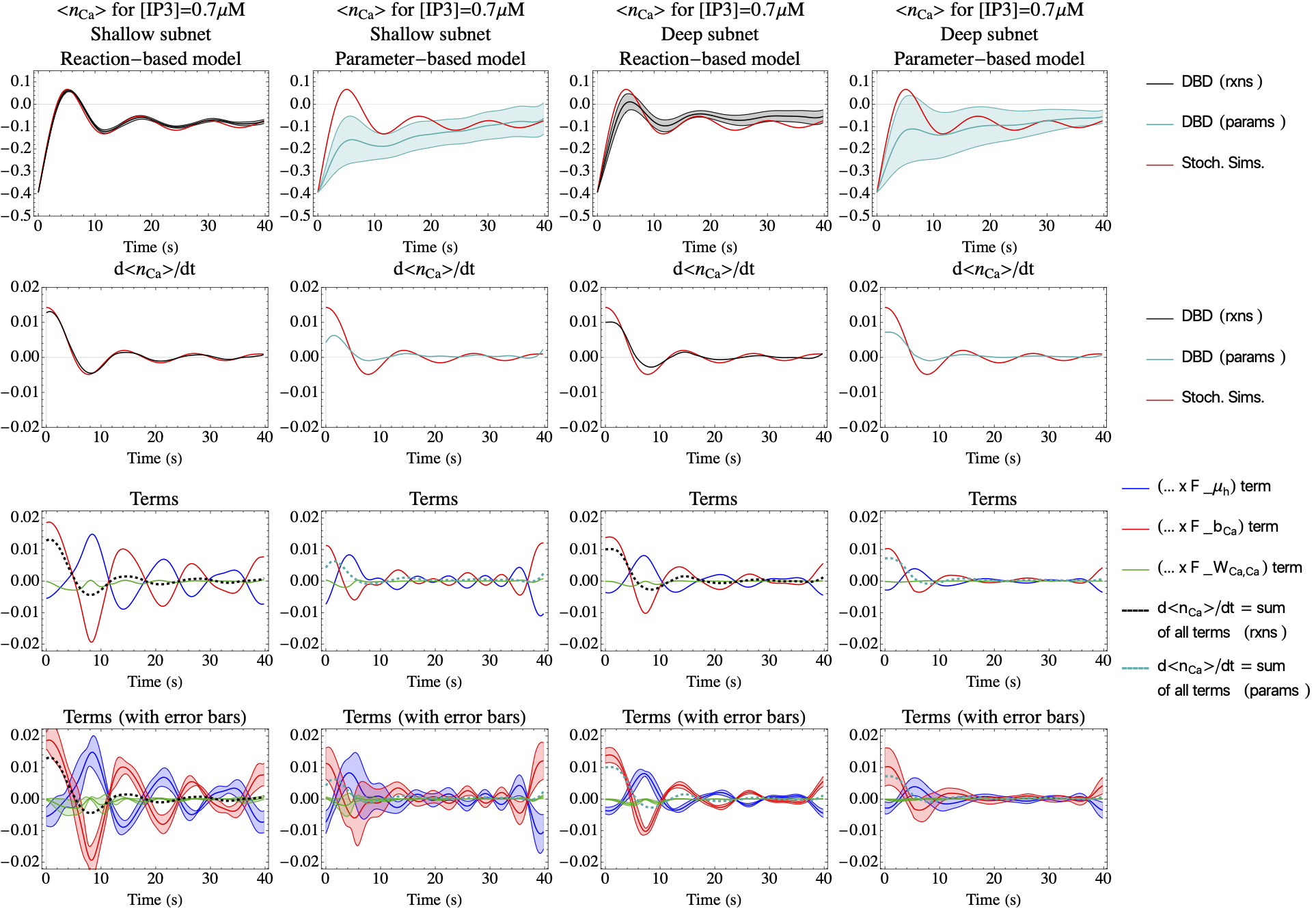}
	\caption{
    Moment closure terms learned corresponding to the model of Figure~3 of the manuscript.
    \textit{First column:} the shallow subnet model for the reaction-based framework.
    \textit{Second column}: shallow subnet, parameter model.
    \textit{Third column}: deep subnet, reaction model.
    \textit{Fourth column}: deep subnet, parameter model.
    \textit{First row:} At a single concentration of $\ipt = 0.7\mu$M, the mean number of calcium $\langle n_\text{Ca} \rangle$ is shown for DBD models, with ground truth from the stochastic simulations in red. 
    \textit{Second row:} The derivative in time of the mean calcium concentration: $d \langle n_\text{Ca} \rangle / dt$. 
    \textit{Third row:} The terms in Equation (3) for the mean calcium $X(\nbf) = n_\text{Ca}$. 
    \textit{Fourth row:} Same as third row with error bars from $10$ optimization trials.
	}
	\label{fig:terms}
\end{figure*}

Figure~\ref{fig:terms} plots the terms for the mean number of calcium $X(\nbf) = n_\text{Ca}$ in the models corresponding to Figure~3 of the manuscript. The terms show that the $F_{\mu_h}$ term is learned to be a counter-phase variable to the $F_{b_\text{Ca}}$ term. The oscillations for the reaction-based model are more coherent, have higher amplitude and frequency on the order of the calcium oscillations. On the other hand, the parameter-based model oscillates at a higher frequency and lower amplitude. This shows that in the comparison model, the inaccuracies in the learned parameters result partially from finding a flawed latent representation in $\mubf_h$ and $\Sigma_h$.

\subsection{Comparison model without reaction approximations}

\begin{figure*}[htp]
	\centering
	\includegraphics[width=0.8\textwidth]{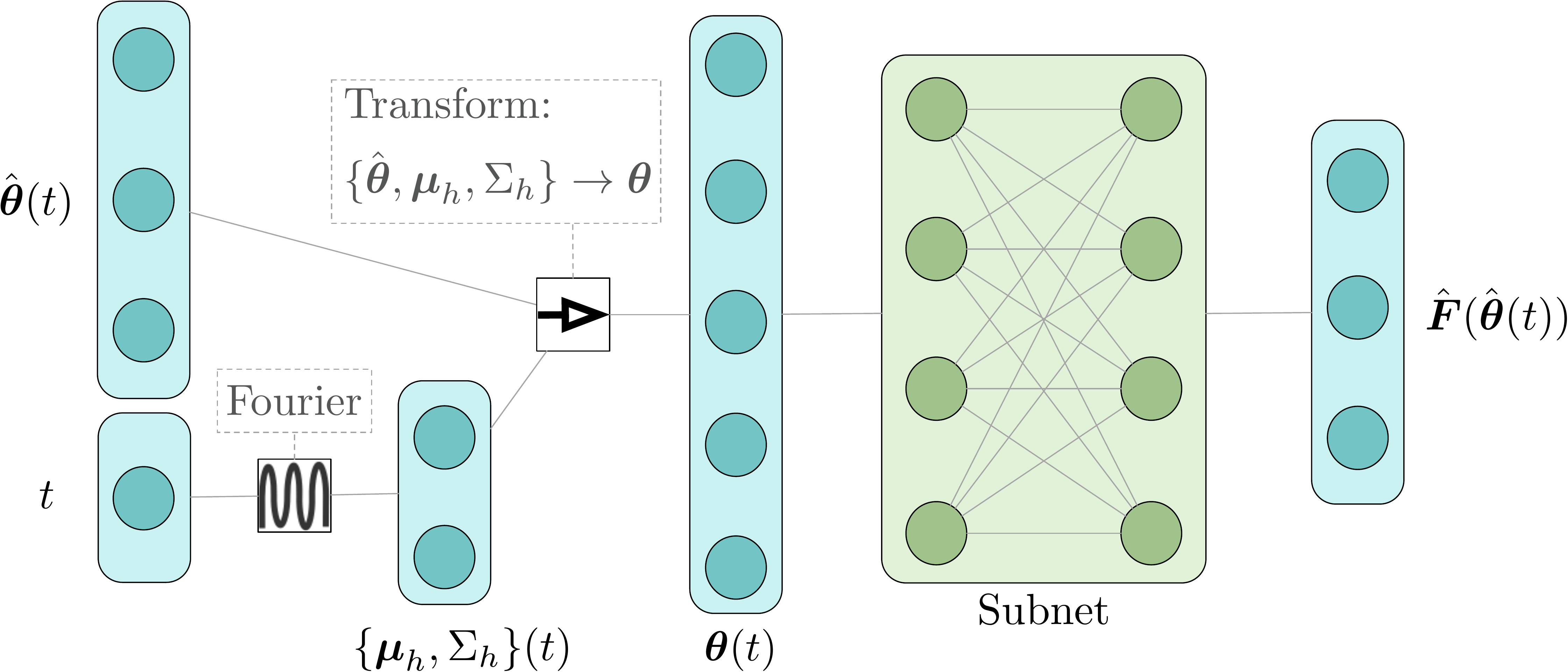}
	\caption{Comparison architecture similar to \figOne, but missing the analytically derived reaction approximations. Instead, the inputs to the subnet model are the interaction parameters directly.}
	\label{fig:compare}
\end{figure*}

Figure~\ref{fig:compare} shows the architecture of the comparison model to \figOne. The network architecture is equivalent except that the reaction approximations are missing. Therefore, the network must learn the functional forms from scratch from the parameters $\thetabf(t)$.

\subsection{Mean-squared error (MSE)}


The mean-squared error (MSE) over parameters in learned models is shown in \figFour. Let $\hat{\thetabf}_\text{int}(t;[\ipt])$ be the integrated parameters after learning the reduced model for a single $\ipt$ concentration, and $\hat{\thetabf}_\text{ML}(t;[\ipt])$ the maximum likelihood parameters identified from the data. The MSE at a single $\ipt$ concentration is then given by:
\begin{equation}
\text{MSE}([\ipt]) = \frac{1}{T} \sum_{t=1}^T |\hat{\thetabf}_\text{int}(t;[\ipt]) - \hat{\thetabf}_\text{ML}(t;[\ipt]) |^2 .
\end{equation}

\end{document}